\documentclass[10pt,twocolumn,letterpaper]{article}

\usepackage[pagenumbers]{cvpr} 

\definecolor{cvprblue}{rgb}{0.21,0.49,0.74}
\usepackage[pagebackref,breaklinks,colorlinks,allcolors=cvprblue]{hyperref}

\usepackage{algorithm}
\usepackage{algpseudocode}
\usepackage{amsmath}  
\usepackage{amssymb}
\usepackage{multirow}
\usepackage{float}
\usepackage{wrapfig}
\usepackage{multibib}
\usepackage{graphicx}
\usepackage{subcaption}
\usepackage{stfloats}

\def\confName{CVPR}
\def\confYear{2027}

\title{RAPID: Codec-Aware Video Decoding for Efficient Long-Video Understanding}

\author{Yuna Lee\\
Dept. of Electrical and Computer Engineering\\
Sungkyunkwan University\\
{\tt\small yuna6548@skku.edu}
\and
Yulhwa Kim\thanks{Corresponding author}\\
Dept. of Semiconductor Systems Engineering\\
Sungkyunkwan University\\
{\tt\small yulhwakim@skku.edu}
}

\begin{document}
\maketitle
\begin{abstract}
Recent vision-language models (VLMs) can process hour-long videos, but the large number of input frames incurs substantial inference overhead. Existing approaches to efficient long-video understanding largely overlook video decoding, which can take longer than LLM prefill. We propose RAPID (Reference- and Access-Point-Informed Decoding), a codec-aware framework that exploits information embedded in compressed bitstreams to reduce both frame decoding and selection costs. RAPID consists of two components. 
RAPID-Dec exploits reference-frame information and random access points to skip unnecessary frames and decode the retained frames in parallel while preserving exact reconstruction of the target frames.
It achieves up to a 3.7$\times$ decoding speedup over the fastest baseline on H.264/AVC videos. RAPID-Sel uses keyframes as coarse representatives of video segments, avoiding the dense candidate decoding and scoring required by existing frame selection methods. It achieves comparable accuracy with up to a 3.0$\times$ preprocessing speedup. Together, RAPID achieves a 4.9$\times$ time to first token (TTFT) speedup on hour-long videos over a state-of-the-art frame selection method using the fastest baseline decoder.
\end{abstract}
\begin{figure*}[!b]
\vspace{-10pt}
  \centering
  \begin{subfigure}{0.32\linewidth}
    \includegraphics[width=\linewidth]{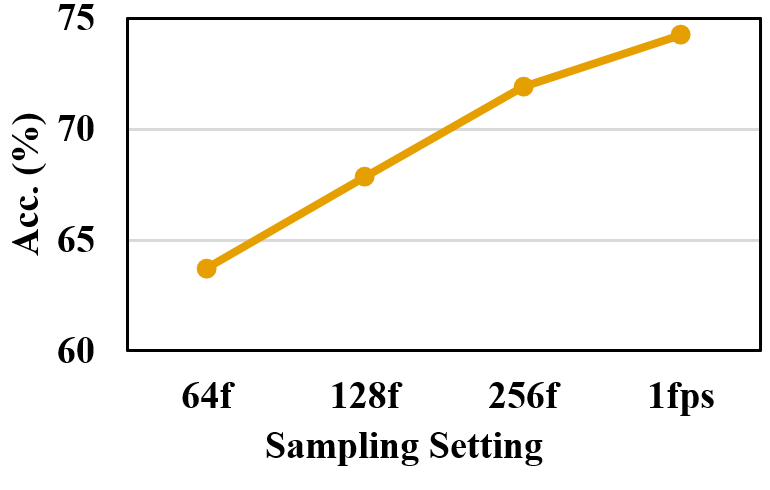}
    \vspace{-17pt}
    \caption{Accuracy vs. sampling setting}
    \label{fig:motivation-a}
  \end{subfigure}
  \hfill
  \begin{subfigure}{0.32\linewidth}
    \includegraphics[width=\linewidth]{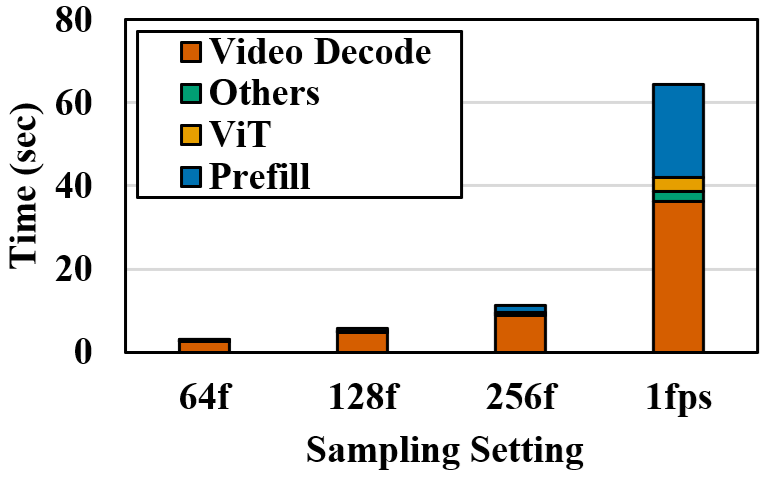}
    \vspace{-17pt}
    \caption{TTFT breakdown}
    \label{fig:motivation-b}
  \end{subfigure}
  \hfill
  \begin{subfigure}{0.32\linewidth}
    \includegraphics[width=\linewidth]{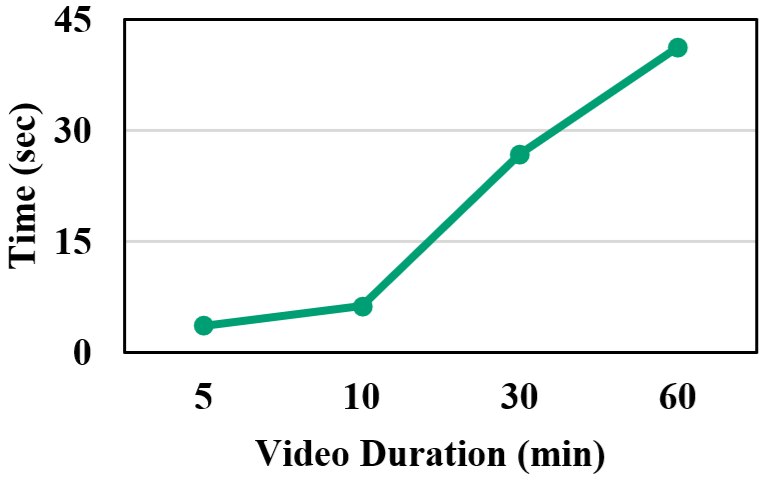}
    \vspace{-17pt}
    \caption{1 fps video decoding latency}
    \label{fig:motivation-c}
  \end{subfigure}
  \vspace{-9pt}
    \caption{
    Motivation of RAPID. (a) Accuracy of Qwen3-VL-8B on MLVU under different frame sampling settings. (b) Breakdown of TTFT for Qwen3-VL-8B across sampling settings. (c) Video decoding latency at 1~fps without frame capping for different video durations.
    In (a) and (b), f denotes the number of uniformly sampled frames (64, 128, or 256), while fps denotes frames per second, with sampling capped at 2,048 frames for the 1~fps setting. All videos were decoded using TorchCodec~\citep{torchcodec}.
    }
  \label{fig:motivation}
\end{figure*}
\section{Introduction}
\label{sec:intro}
Vision-language models (VLMs) have advanced rapidly, and recent models extend their scope from images and short clips to long videos~\citep{Qwen3-VL, Qwen2_5_VL, glm46, wang2025internvl3_5}. As their context length grows, these models can take more frames as input, improving their accuracy on long videos, as shown in Fig.~\ref{fig:motivation-a}. However, processing long videos incurs substantial inference overhead, including video decoding, which converts compressed bitstreams into RGB frames.


Fig.~\ref{fig:motivation-b} shows the time to first token (TTFT) breakdown of Qwen3-VL-8B on hour-long videos under different frame sampling settings. In every setting, video decoding takes even longer than LLM prefill, as also reported by a recent study~\citep{beforefirsttoken}. Moreover, existing frame selection methods decode and score dense candidate frames, typically sampled at 1 frame per second (fps)~\cite{aks, wfs_bs}, before selecting a small subset for the VLM. This further increases preprocessing latency, as shown in Fig.~\ref{fig:motivation-c}. Existing video decoding accelerators parallelize decoding across CPU cores or GPUs~\cite{schneider2026quickvideo,zhao2025flashcodec}, but still reconstruct frames that the VLM never uses. Therefore, efficient long-video understanding requires reducing decoding costs both for obtaining the frames fed to the VLM and for selecting those frames.

We observe that compressed video bitstreams contain useful information about frame dependencies and random access points, which can be obtained without decoding the frames. 
In addition, video encoders can place keyframes at scene changes to improve compression efficiency~\citep{x264, x265}.
These properties provide opportunities to accelerate decoding and frame selection without modifying the VLM. Based on these observations, we propose RAPID, a codec-aware framework consisting of RAPID-Dec for decoding target frames and RAPID-Sel for selecting them.

Our contributions are summarized as follows:
\begin{itemize}
\item 
We propose RAPID-Dec, which leverages information present in the bitstream to skip unnecessary frames and decode the retained frames in parallel without approximation,
achieving up to a 3.7$\times$ speedup in video decoding over the fastest baseline decoder on H.264/AVC videos.
\item 
We propose RAPID-Sel, which uses keyframes as coarse representatives of video segments to select query-relevant frames. By decoding and scoring only keyframes instead of dense candidate frames, RAPID-Sel achieves comparable accuracy with up to a 3.0$\times$ speedup in preprocessing.
\item We combine the two components into RAPID, which achieves a 4.9$\times$ speedup in TTFT on hour-long videos over a state-of-the-art frame selection method with the fastest baseline decoder.
\end{itemize}
\section{Related work} \label{sec:related}

\subsection{Efficient video decoding}
QuickCodec~\citep{schneider2026quickvideo} and FlashCodec~\citep{zhao2025flashcodec} accelerate video decoding for VLMs by partitioning a video into segments and decoding them in parallel. QuickCodec assigns each segment to a CPU core, while FlashCodec distributes segments across GPUs and their NVDEC hardware decoders. However, both methods decode frames that are neither sampled as VLM input nor referenced by other frames. QuickCodec decodes every frame regardless of the sampling interval, limiting its benefit to dense sampling. FlashCodec is constrained by the limited number of NVDEC engines per GPU, and its reference frames occupy GPU memory shared with the model.
In video analytics, Smol~\citep{smol} decodes low-resolution inputs and fine-tunes DNNs on them, while CoVA~\citep{hwang2022cova} trains a lightweight network on codec metadata to select frames for object detection. However, both methods rely on approximations or task-specific training. In contrast, RAPID-Dec skips unnecessary frames while preserving exact reconstruction, without task-specific training.

\subsection{Efficient long-video understanding}

\noindent\textbf{Frame selection.}
Under a limited frame budget, frame selection seeks higher accuracy than uniform sampling.
For action recognition, SCSampler~\citep{korbar2019scsampler} trains a lightweight sampler on compressed-domain features to select salient clips.
For VLMs, \citet{mllm_select_hu2025m} and \citet{rl_frame_select_qin2026efficient} train a lightweight external selector to find query-relevant frames, and Frame-Voyager~\citep{voyager_yu2025frame} trains a plug-in module attached to the VLM.
However, these methods require video training data with supervision for frame selection.
In contrast, training-free methods score frames with a pretrained lightweight vision-language model~\citep{li2022blip, clip}.
AKS~\citep{aks} and WFS-SB~\citep{wfs_bs} score candidate frames sampled at 1~fps. AKS balances query relevance and temporal coverage, whereas WFS-SB allocates the frame budget across semantic segments.
Focus~\citep{ziruiz2026focus} first identifies query-relevant temporal regions and then selects the top-scoring frames within them.
However, these methods must decode and score a pool of candidate frames much larger than the final frame budget, introducing substantial preprocessing overhead.

\medskip
\noindent\textbf{Token compression and codec-based representations.}
Token compression methods reduce visual tokens within or across frames~\citep{bolya2022tome, yang2025visionzip, tao2025dycoke, tang2026surge, zou2026codecsight, shao2025holitom, wang2026earlytom}, but operate on decoded frames and therefore do not reduce video decoding latency. Other approaches exploit compressed-domain representations to reduce reliance on RGB frames. CoPE-VideoLM~\citep{sarkar2026copevideolm} and ReMoRa~\citep{yashima2026remora} replace most RGB frames with codec-derived information, such as motion vectors and residuals, while CoViAR~\citep{coviar} explores a similar approach for action recognition. Unlike RAPID, these approaches require training on compressed-domain inputs.



\section{Method} \label{sec:method}

\subsection{Preliminaries: video codec}\label{sec:codec-explain}
A video codec consists of an encoder and a decoder. The encoder compresses a video into a compact bitstream whose syntax is defined by a coding standard such as H.264/AVC~\citep{avc_overview} or H.265/HEVC~\citep{hevc_overview}, and the decoder reconstructs the frames from the bitstream when the video is presented to users or VLMs. For VLMs, the reconstructed frames are further converted into RGB images. Existing works exploit compressed-domain cues that reflect the visual content, such as motion vectors and residuals, for efficient video understanding~\cite{zou2026codecsight, sarkar2026copevideolm, yashima2026remora}. 
Beyond these content cues, we exploit the bitstream syntax, which reveals frame dependencies and random access points, to skip unnecessary decoding and guide frame selection.

\begin{figure}[t]
  \centering
  \begin{subfigure}{\linewidth}
    \centering
    \includegraphics[width=\linewidth]{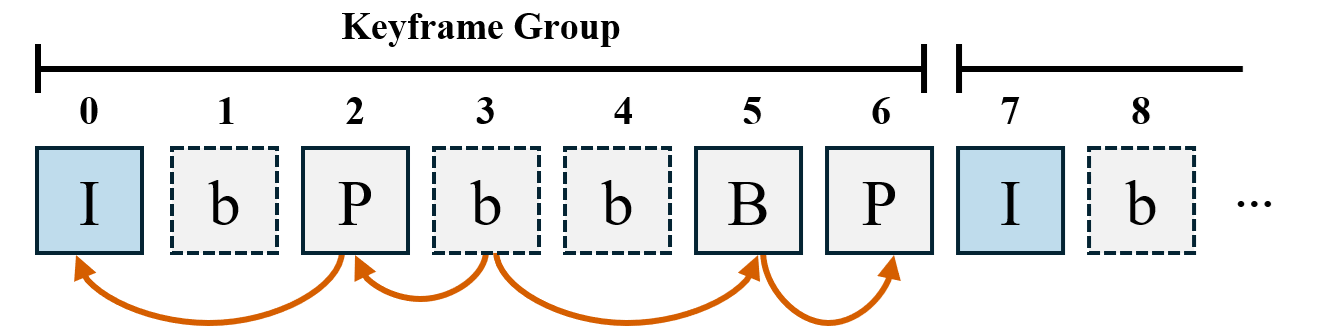}
    \caption{Frames in display order}
    \label{fig:codec-gop}
  \end{subfigure}

  \vspace{2mm}
  \begin{subfigure}{\linewidth}
    \centering
    \includegraphics[width=\linewidth]{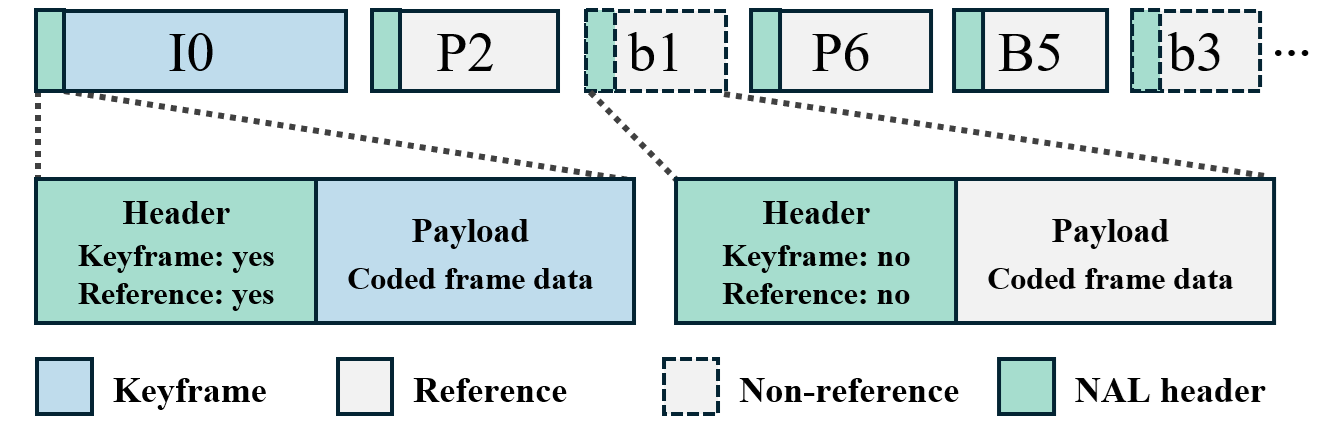}
    \caption{Bitstream in decoding order}
    \label{fig:codec-nal}
  \end{subfigure}
  \vspace{-20pt}
    \caption{Example of keyframe groups. (a) In display order, each arrow points from a frame to a frame it references. (b) In the bitstream, frames are stored in decoding order, and the bitstream consists of NAL units, each with a header and a payload.}
  \label{fig:codec}
\end{figure}

\medskip
\noindent\textbf{Frame types in codec.}
Depending on the prediction method used for encoding, frames are classified into I-, \mbox{P-,} and B-frames.
An I-frame is encoded using only intra prediction and thus can be decoded independently of other frames.
In contrast, P- and B-frames additionally utilize inter prediction. A P-frame is predicted from previously decoded frames, whereas a B-frame can be bi-directionally predicted from both preceding and subsequent frames in display order. 
As shown in Fig.~\ref{fig:codec-gop}, frame 0, an I-frame, can be decoded independently, but frame 3 can be decoded only after frames 2 and 5 are decoded. Due to this dependency, obtaining even a single frame requires decoding all the frames it depends on, which makes video decoding a bottleneck in VLM pipelines that sparsely sample frames from long videos.

Some I-frames further serve as random access points (RAPs), from which the decoder can start decoding without any preceding frames.
Although they are named differently, RAPs are provided in all major video codecs to support functionalities such as seeking~\citep{avc_overview, hevc_overview, mukherjee2013vp9, wang2021vvc_hls, av1}.
H.264/AVC defines the instantaneous decoding refresh (IDR) frame~\citep{avc_overview}, and H.265/HEVC defines intra random access point frames, comprising IDR, clean random access, and broken link access frames~\citep{hevc_overview}.
Throughout the paper, we refer to a RAP as a \emph{keyframe} if no subsequent frame in the bitstream references any frame preceding it.
Frames after a scene change share little content with the frames before it, so inter prediction across a scene change is inefficient~\citep{adpativeiframe, sastre2006shotiframe}. Accordingly, widely used encoders~\citep{x264, x265} insert keyframes at detected scene changes by default.

As RAPID accesses the bitstream at keyframe granularity, we define a \emph{keyframe group} as a sequence of frames from a keyframe up to the frame before the next keyframe, as illustrated in Fig.~\ref{fig:codec-gop}. 
Each keyframe group starts from a keyframe and can thus be decoded independently.

\medskip
\noindent\textbf{Syntax structure of bitstream.} 
The bitstreams of standard codecs such as H.264/AVC and H.265/HEVC are composed of a sequence of network abstraction layer (NAL) units, as shown in Fig.~\ref{fig:codec-nal}. Each NAL unit consists of a header and a payload. The header indicates the type of the payload as well as frame-level properties, such as whether the frame is a keyframe and whether it can be referenced by other frames, while the payload carries the coded data. Since the header is a fixed-length field located at the beginning of each NAL unit, this information can be obtained by parsing only a few bytes, without decoding the frame.

\begin{figure*}
  \centering
  \includegraphics[width=\linewidth]{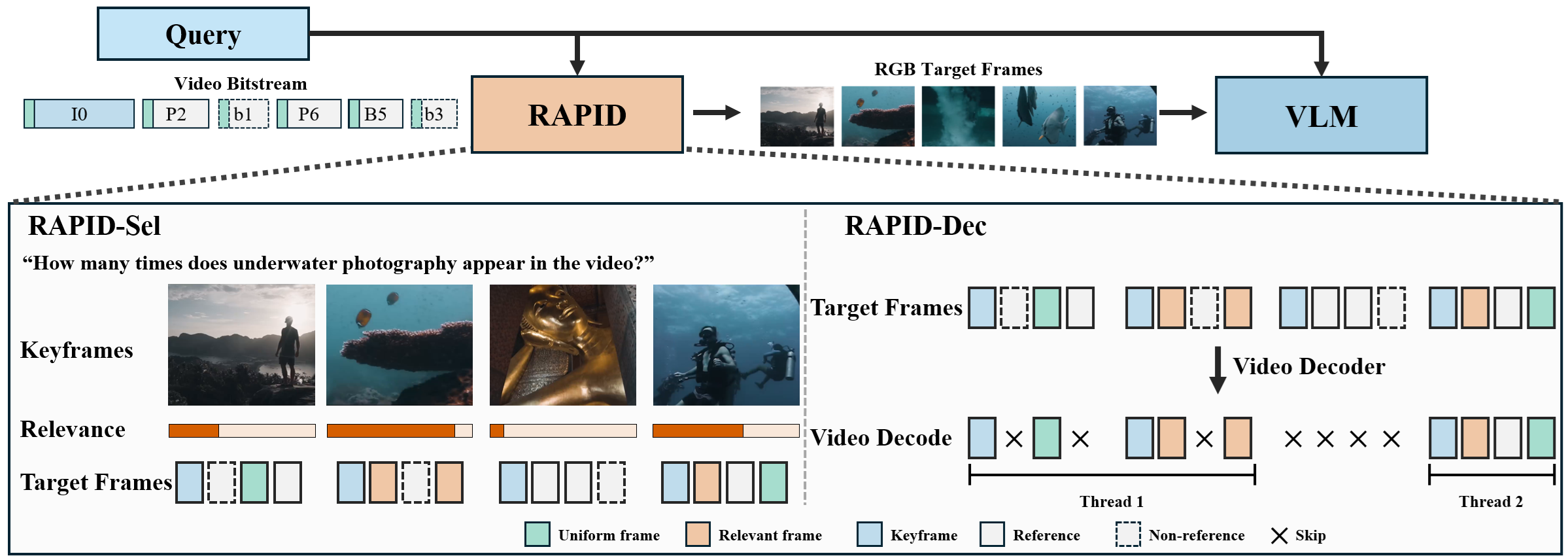}
  \vspace{-18pt}
    \caption{
    Overview of RAPID. RAPID-Dec (right) decodes target frames chosen by any sampling strategy, including RAPID-Sel (left). It skips groups without target frames, frames after the last target frame of each group, and non-reference frames other than target frames, while preserving exact reconstruction of the targets. The retained groups are then distributed across threads for parallel decoding. RAPID-Sel decodes and scores only the keyframes against the query and selects target frames by combining uniform frames with relevant frames allocated to groups by relevance.
    }
  \label{fig:overview}
  \vspace{-10pt}
\end{figure*}
\subsection{Overview of RAPID}
RAPID consists of two components, RAPID-Dec and RAPID-Sel, as illustrated in Fig.~\ref{fig:overview}.
RAPID-Dec accelerates video decoding for any frame sampling strategy by filtering out unnecessary frames from the bitstream before decoding the retained frames.
The target frames can be chosen by uniform sampling, existing frame selection methods, or RAPID-Sel.
RAPID-Sel identifies query-relevant scenes at a coarse level using keyframes, selecting more query-relevant target frames than uniform sampling without decoding dense candidate frames.

\subsection{RAPID-Dec}\label{sec:rapid-dec}
A decoder for playback must reconstruct every frame, because every frame is displayed.
A VLM, in contrast, consumes only the frames chosen by its sampling strategy.
The other frames are reconstructed only to be discarded, or to serve as references for the chosen ones.
The video decoding latency of a VLM pipeline is therefore governed not by how many frames the model uses, but by how many frames the decoder must reconstruct to obtain them.
RAPID-Dec narrows this gap using the syntax of the bitstream.
Given the indices of the target frames, RAPID-Dec determines which frames to retain for exact reconstruction, delivers only their bitstream to the decoder, and decodes them in parallel. The decoded targets are identical to those obtained by decoding the entire video.

\medskip
\noindent\textbf{Retained frames.}
RAPID-Dec first scans the bitstream once without decoding any frame, parsing only a few short syntax elements of each frame.
This scan yields the locations of the keyframes, and hence the groups, together with the reference flag of every frame. We call frames that cannot be referenced non-reference frames.

Let $\mathcal{T}$ be the set of target frames, and $G_g$ the $g$-th group with keyframe $k_g$.
For each group that contains at least one target, let $\ell_g$ be its last target in decoding order.
RAPID-Dec defines the frames to retain in $G_g$ as
\begin{equation}
\mathcal{E}_g = \{\, f \in G_g : f \preceq \ell_g,\ \mathrm{ref}(f) \lor f \in \mathcal{T} \cup \{k_g\} \,\},
\label{eq:essential}
\end{equation}
where $f \preceq \ell_g$ means that $f$ precedes or equals $\ell_g$ in decoding order, and $\mathrm{ref}(f)$ is the reference flag in the NAL unit header of $f$.
Accordingly, three kinds of frames are removed from the bitstream before it reaches the decoder:
(i) all frames of groups without targets,
(ii) frames after the last target of each group in decoding order, which no earlier frame references since every frame is stored after all the frames it references, and
(iii) non-reference frames that are not targets.
Since none of the remaining frames references a removed frame, every target is reconstructed exactly as in full video decoding, without approximation.
Note that (ii) relies on decoding order rather than display order, because a frame may reference a frame displayed after it. For example, frame 3 in Fig.~\ref{fig:codec-gop} references frame 5, which in turn references frame 6. Both are displayed after frame 3 but stored before it (Fig.~\ref{fig:codec-nal}).
By conservatively retaining potentially required reference frames, RAPID-Dec avoids heavy explicit dependency tracing while guaranteeing exact reconstruction of the target frames.


The savings from (i) and (ii) grow as sampling becomes sparser, whereas (iii) applies even to dense sampling, where almost every group contains a target.
To estimate the room for (iii), we measure the ratio of non-reference frames in each video of Video-MME~\citep{videomme}, LongVideoBench~\citep{wu2024longvideobench}, and MLVU~\citep{zhou2025mlvu}.
As shown in Table~\ref{tab:nonref_ratio}, non-reference frames account for 37.5--52.7\% of the frames on average, which indicates a large potential saving from skipping them.

FFmpeg~\citep{ffmpeg} can also skip non-reference frames with its \texttt{skip\_frame} option, but it cannot select which ones to skip and thus also skips non-reference targets. RAPID-Dec instead removes only the non-reference frames that are not targets, as in Eq.~(\ref{eq:essential}).

\begin{table}
  \centering
  \caption{Average ratio of non-reference frames per video.}
  \vspace{-8pt}
  \label{tab:nonref_ratio}
  \begin{tabular}{@{}ccc@{}}
    \toprule
    Video-MME & LongVideoBench & MLVU \\
    \midrule
    52.7\% & 51.8\% & 37.5\% \\
    \bottomrule
  \end{tabular}
  \vspace{-12pt}
\end{table}

\medskip
\noindent\textbf{Workload-balanced parallel decoding.}
The filtered keyframe groups are then decoded in parallel. Given a thread budget $T$, let $\mathcal{G}$ be the set of groups that contain targets, and $N = \min(T, |\mathcal{G}|)$. 
A simple way to parallelize decoding is to split $\mathcal{G}$ into $N$ contiguous segments with the same number of groups and let each worker decode its segment. However, groups differ in decoding cost. Keyframe groups can differ in length, and once the frames after the last target are removed, the cost of a group further depends on where its last target lies. Segments with the same number of groups can therefore leave workers unevenly loaded, and the slowest worker determines the latency.
RAPID-Dec instead measures the workload of each group by the number of its retained frames, $w_g = |\mathcal{E}_g|$.
It splits $\mathcal{G}$ into at most $N$ contiguous segments of approximately equal total workload, cutting where the cumulative workload reaches multiples of $\frac{1}{N}\sum_{g \in \mathcal{G}} w_g$, and assigns each segment to a worker.
When fewer than $T$ workers are created because only a few groups contain targets, the remaining threads are used by the multithreading inside each decoder, so the total number of threads stays within $T$.

Each worker owns a demuxer and a decoder.
It visits the groups of its segment in order, seeking directly to each keyframe and feeding only the retained frames to its decoder (Algorithm~\ref{alg:rapid-dec}).
Each worker converts only its decoded targets to RGB and writes them directly to their positions in the output, so the post-processing is parallelized as well.

\begin{algorithm}[t]
\caption{RAPID-Dec}\label{alg:rapid-dec}
\begin{algorithmic}[1]
\Require bitstream $\mathcal{B}$, target frames $\mathcal{T}$, thread budget $T$
\Ensure target frames in RGB $\{x_t\}_{t \in \mathcal{T}}$
\State Scan $\mathcal{B}$ without decoding $\rightarrow$ keyframe groups $\{G_g\}$ with keyframes $k_g$, reference flags
\State $\mathcal{G} \gets \{g : G_g \cap \mathcal{T} \neq \emptyset\}$, in order
\ForAll{$g \in \mathcal{G}$}
  \State $\ell_g \gets$ last target of $G_g$ in decoding order
  \State $\mathcal{E}_g \gets$ Eq.~\eqref{eq:essential}
  \State $w_g \gets |\mathcal{E}_g|$
\EndFor
\State $N \gets \min(T, |\mathcal{G}|)$
\State Split $\mathcal{G}$ into $M \le N$ contiguous segments $\{S_i\}$ with $\sum_{g \in S_i} w_g \approx \frac{1}{N}\sum_{g \in \mathcal{G}} w_g$
\ForAll{$S_i$ \textbf{in parallel}}
  \State Open a decoder with $\lfloor T/M \rfloor$ threads
  \ForAll{$g \in S_i$}
    \State Seek to $k_g$ and decode $\mathcal{E}_g$
    \State Write the decoded targets in RGB to the output
  \EndFor
\EndFor
\end{algorithmic}
\end{algorithm}

\subsection{RAPID-Sel}
\label{sec:rapid-sel}
RAPID-Dec accelerates video decoding for any target frames, but the accuracy of the VLM depends on which frames are chosen.
Without frame selection, the target frames are typically sampled uniformly, which may miss query-relevant scenes when the frame budget cannot cover a long video.
Query-aware frame selection methods~\citep{aks, wfs_bs} instead decode dense candidate frames, typically at 1~fps, and score them against the query with a lightweight vision-language model~\citep{clip, li2022blip} to select the target frames.
However, decoding the candidates, which far outnumber the frame budget in long videos, makes selection slow.
RAPID-Sel aims to improve accuracy over uniform sampling with minimal latency overhead.

Since keyframes are typically inserted at scene changes (Sec.~\ref{sec:codec-explain}), the keyframe $k_g$ can serve as a coarse representative of the content of $G_g$.
Moreover, keyframes reference no other frames, so they can all be decoded in parallel.
RAPID-Sel therefore decodes only the keyframes of the $K$ groups and measures the relevance of each group $G_g$ to a text query $q$ by the matching logit of an image-text scorer such as CLIP,
\begin{equation}
s_g = \tau \cos\big(\phi_I(k_g),\, \phi_T(q)\big),
\label{eq:logit}
\end{equation}
where $\phi_I$ and $\phi_T$ are the image and text encoders of the scorer and $\tau$ is its logit scale.

Given a frame budget $B$ and a ratio $\rho \in [0, 1]$, RAPID-Sel divides it into a uniform budget $B_u = \mathrm{round}(\rho B)$ and a relevance budget $B_r = B - B_u$.
The $B_u$ uniform frames, sampled evenly over the entire video, preserve temporal coverage for queries about the whole video.
The $B_r$ relevant frames are assigned one at a time according to the probabilities over groups given by the logits, each to the group $g^\ast$:
\begin{equation}
p_g = \frac{e^{s_g}}{\sum_{h=1}^{K} e^{s_h}}, \qquad
g^\ast = \arg\max_{g} \frac{p_g}{c_g + 1},
\label{eq:alloc}
\end{equation}
where $p_g$ is the probability of $k_g$ matching $q$ among the $K$ keyframes, and $c_g$ is the number of relevant frames already assigned to $G_g$.
This rule, known as the D'Hondt method~\citep{balinski1978jefferson_dhondt}, allocates frames approximately in proportion to $p_g$.
Since the probability of a group is discounted by the frames it already holds, frames concentrate on one group only as far as its probability dominates, and otherwise spread over several groups.
A group is full once it holds one selected frame per second of its duration, since frames less than a second apart are likely redundant.
Full groups are excluded from the $\arg\max$ in Eq.~\eqref{eq:alloc}.

\begin{figure*}[t]
  \centering
  \begin{subfigure}{0.32\linewidth}
    \includegraphics[width=\linewidth]{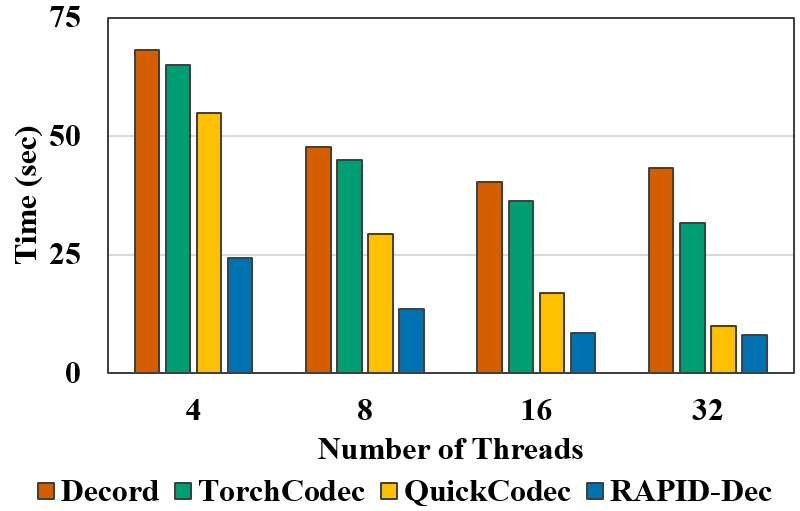}
    \caption{H.264/AVC}
    \label{fig:parallel_lat-a}
  \end{subfigure}
  \hfill
  \begin{subfigure}{0.32\linewidth}
    \includegraphics[width=\linewidth]{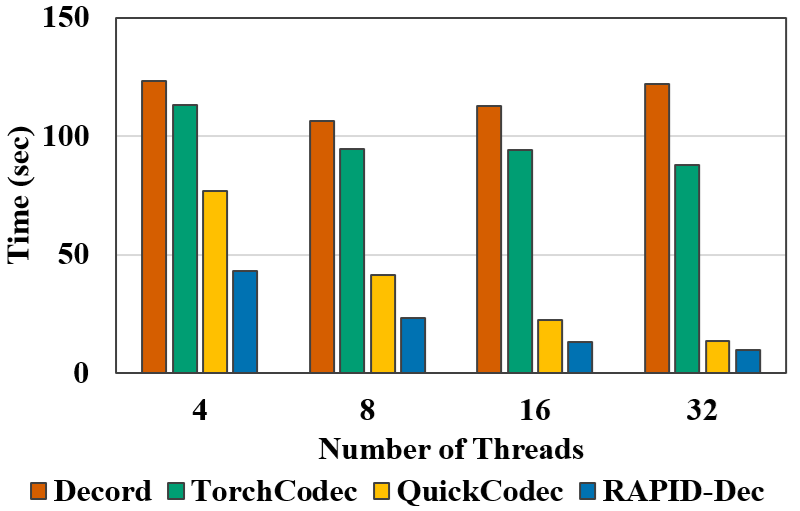}
    \caption{H.265/HEVC}
    \label{fig:parallel_lat-b}
  \end{subfigure}
  \hfill
  \begin{subfigure}{0.32\linewidth}
    \includegraphics[width=\linewidth]{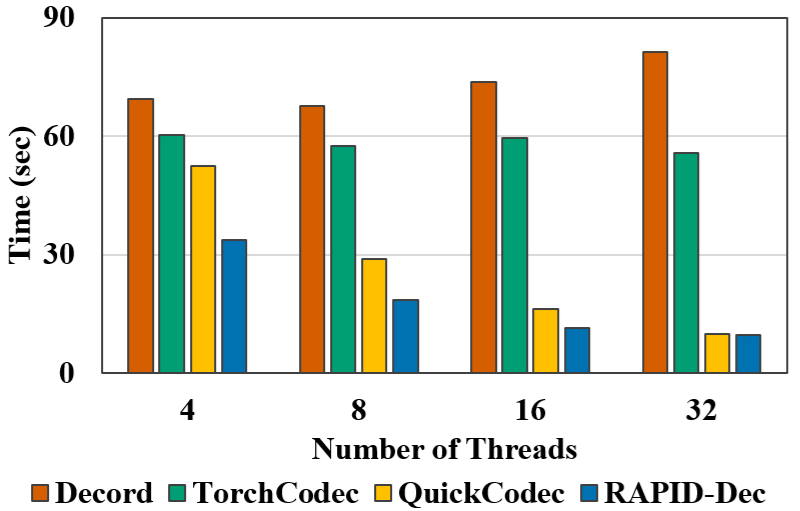}
    \caption{VP9}
    \label{fig:parallel_lat-c}
  \end{subfigure}
  \vspace{-8pt}
  \caption{Video decoding latency of 1~fps sampling (up to 2,048 frames) with different numbers of threads.}
  \label{fig:parallel_lat}
\end{figure*}

\begin{figure*}[t]
  \centering
  \begin{subfigure}{0.32\linewidth}
    \includegraphics[width=\linewidth]{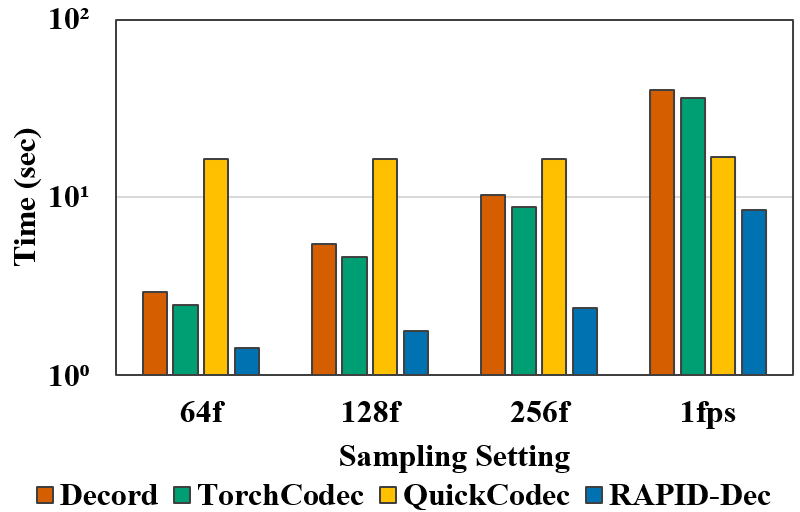}
    \caption{H.264/AVC}
    \label{fig:sample_lat-a}
  \end{subfigure}
  \hfill
  \begin{subfigure}{0.32\linewidth}
    \includegraphics[width=\linewidth]{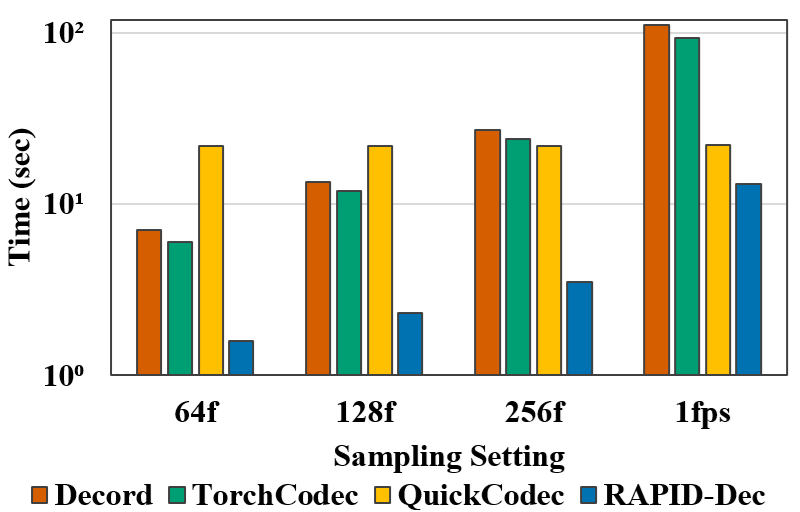}
    \caption{H.265/HEVC}
    \label{fig:sample_lat-b}
  \end{subfigure}
  \hfill
  \begin{subfigure}{0.32\linewidth}
    \includegraphics[width=\linewidth]{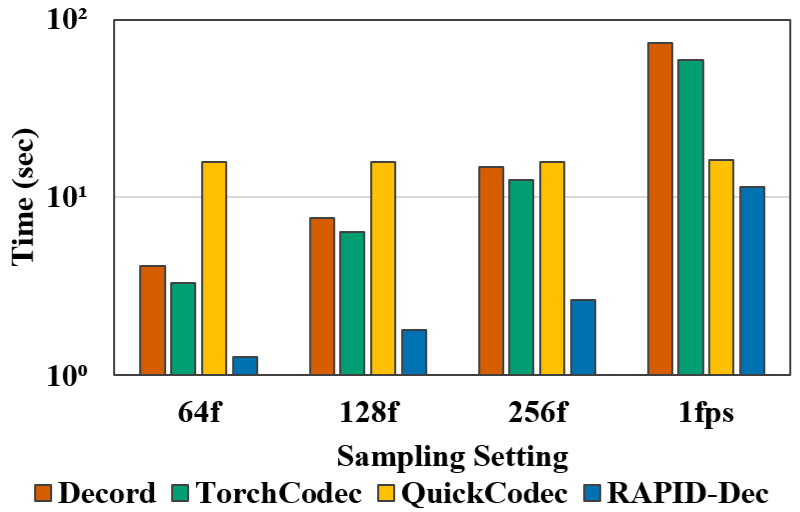}
    \caption{VP9}
    \label{fig:sample_lat-c}
  \end{subfigure}
  \vspace{-8pt}
  \caption{Video decoding latency under different frame sampling settings with 16 threads. The y-axis is in log scale.}
  \label{fig:sample_lat}
\end{figure*}

Finally, the $c_g$ relevant frames are placed evenly within each group by minimizing the sum of squared gaps between consecutive selected frames.
The selected frames form the target set $\mathcal{T}$ of RAPID-Dec, and keyframes among them are reused from the scoring step.
Without RAPID-Sel, $\mathcal{T}$ is given by uniform sampling or an existing frame selection method.


\section{Experiments}
\label{sec:exp}
\subsection{Experimental setup}
\noindent\textbf{Environments.}
We measure the video decoding latency on an Intel Xeon Gold 6530 CPU with its clock frequency fixed at 2.1~GHz, and run the VLM and the CLIP scorer on an NVIDIA RTX 5880 Ada GPU.
The thread budget is set to 16 unless otherwise specified, and every decoder is given the same number of threads.
The reported latency covers the entire process from opening the video file to obtaining the target frames in RGB, including the bitstream scan of RAPID-Dec.
When frames are sampled at 1~fps as the input of the VLM, we cap the number of frames at 2,048 following the evaluation setting of Qwen3-VL~\citep{Qwen3-VL}, whereas candidate frames for frame selection are decoded at 1~fps without the cap.
All accuracy evaluations are conducted with the lmms-eval framework~\citep{lmms_eval2024}.
The ratio $\rho$ of RAPID-Sel is set to 0.5 by default.

\medskip
\noindent\textbf{Baselines.}
For video decoding, we compare RAPID-Dec with three decoding backends: Decord~\citep{decord2019}, TorchCodec~\citep{torchcodec}, and QuickCodec~\citep{schneider2026quickvideo}.
Decord and TorchCodec are widely used libraries for loading videos in VLM evaluation, whereas QuickCodec is designed to accelerate dense frame sampling.

For frame selection, we compare RAPID-Sel with uniform sampling, Top-K selection, AKS~\citep{aks}, Focus~\citep{ziruiz2026focus}, and WFS-SB~\citep{wfs_bs}.
Top-K, AKS, and WFS-SB score candidate frames sampled at 1~fps, and Top-K simply selects the frames with the highest relevance scores.
Focus instead adopts a two-stage coarse-to-fine selection.

\medskip
\noindent\textbf{Models and benchmarks.}
We evaluate Qwen3-VL-8B~\citep{Qwen3-VL} and GLM-4.6V-Flash~\citep{glm46} on three long-video understanding benchmarks: Video-MME~\citep{videomme}, LongVideoBench (LVB)~\citep{wu2024longvideobench}, and MLVU~\citep{zhou2025mlvu}, using Video-MME without subtitles, the validation set of LongVideoBench, and the dev set of MLVU.
We use CLIP ViT-B/32~\citep{clip} to compute the relevance between the query and the candidate frames for every frame selection method.

\subsection{Video decoding efficiency of RAPID-Dec}\label{sec:exp-dec}
We first evaluate RAPID-Dec alone, with target frames given by uniform sampling at various budgets, to isolate video decoding from the choice of target frames.

\medskip
\noindent\textbf{Impact of parallelism.}
Fig.~\ref{fig:parallel_lat} shows the video decoding latency of 1~fps sampling with different numbers of threads on hour-long videos from Video-MME~\citep{videomme}.
To evaluate the generality of RAPID-Dec, we extend the evaluation to H.265/HEVC, a more recent standard codec, and VP9~\citep{mukherjee2013vp9}, an open-source video codec, both re-encoded from the original H.264/AVC videos.
Each result is averaged over four hour-long videos with five runs per video.
RAPID-Dec achieves the lowest latency for every codec and number of threads.
With the default 16 threads, RAPID-Dec is 4.3$\times$, 7.2$\times$, and 5.2$\times$ faster than TorchCodec, the faster of the two widely used libraries, and 2.0$\times$, 1.7$\times$, and 1.4$\times$ faster than QuickCodec on H.264/AVC, H.265/HEVC, and VP9, respectively.
Decord and TorchCodec scale poorly, possibly because their threads only parallelize the decoding inside a single decoder, where inter-frame dependencies might limit the parallelism.

\medskip
\noindent\textbf{Impact of frame sampling policy.}
Fig.~\ref{fig:sample_lat} varies the sampling setting on the same videos.
RAPID-Dec again achieves the lowest latency in every setting for all three codecs.
QuickCodec decodes every frame, so its latency remains nearly constant regardless of the number of sampled frames.
It is therefore effective under dense sampling, but its gap to RAPID-Dec widens as sampling becomes sparser.
The latency of the other decoders grows with the number of sampled frames, but that of RAPID-Dec grows more slowly.
RAPID-Dec's speedup over TorchCodec thus increases with denser sampling, from 1.7$\times$ to 4.3$\times$ on H.264/AVC, from 3.8$\times$ to 7.2$\times$ on H.265/HEVC, and from 2.6$\times$ to 5.2$\times$ on VP9.
Over the fastest baseline in each setting, the speedups reach up to 3.7$\times$, 6.3$\times$, and 4.7$\times$ on H.264/AVC, H.265/HEVC, and VP9, respectively.

\begin{table}[!b]
  \centering
  \caption{Ablation of RAPID-Dec. Decoding latency (Lat.) on videos of Fig.~\ref{fig:sample_lat-a}. Components are added cumulatively, and speedup is relative to the previous row.
  }
  \vspace{-8pt}
  \small
  \setlength{\tabcolsep}{3pt}
  \begin{tabular*}{\linewidth}{@{\extracolsep{\fill}} l rr rr @{}}
    \toprule
    & \multicolumn{2}{c}{64 frames} & \multicolumn{2}{c}{1~fps} \\
    \cmidrule(lr){2-3} \cmidrule(lr){4-5}
    Configuration & Lat. (s) & Speedup & Lat. (s) & Speedup \\
    \midrule
    Parallel decoding only & 15.91 & -- & 16.38 & -- \\
    \;+ Group skipping     & 2.20 & 7.23$\times$ & 15.94 & 1.03$\times$ \\
    \;+ Truncation         & 1.84 & 1.20$\times$ & 13.03 & 1.22$\times$ \\
    \;+ Workload balancing & 1.77 & 1.04$\times$ & 12.90 & 1.01$\times$ \\
    \;+ Non-ref.\ skipping (full) & \textbf{1.45} & 1.22$\times$ & \textbf{8.72} & 1.48$\times$ \\
    \bottomrule
  \end{tabular*}
  \label{tab:dec_ablation}
\end{table}
\medskip
\noindent\textbf{Impact of each component.}
We analyze the contribution of each component in the setting of Fig.~\ref{fig:sample_lat-a} at 64 frames and 1~fps sampling in Table~\ref{tab:dec_ablation}.
We start from parallel decoding alone, which decodes every frame in contiguous segments with the same number of groups, and add the components of RAPID-Dec.
The most effective component depends on the sampling setting.
With 64 frames, at most 64 groups contain targets, so group skipping yields the largest speedup (7.23$\times$).
At 1~fps, nearly every group contains a target, and non-reference skipping instead yields the largest speedup (1.48$\times$) as the only component that removes frames before the last target of each group.
Thus, group skipping is most effective for sparse sampling and non-reference skipping for dense sampling, but every component reduces the latency in both settings, making the full RAPID-Dec 11.0$\times$ and 1.88$\times$ faster than parallel decoding alone with 64 frames and 1~fps, respectively.

\begin{figure}[t]
  \centering
  \includegraphics[width=\linewidth]{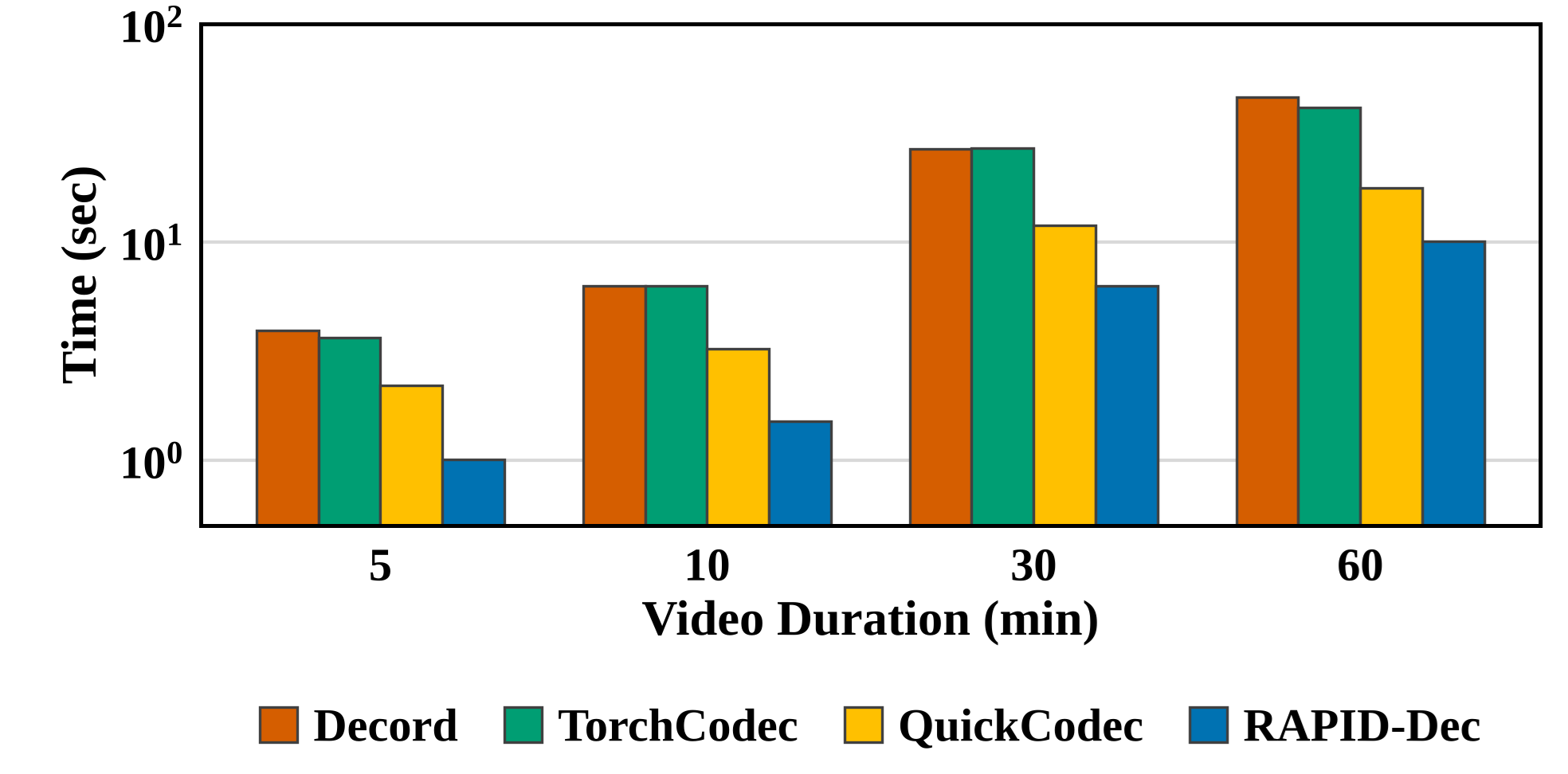}
  \vspace{-18pt}
  \caption{Latency of decoding frames sampled at 1~fps from H.264/AVC videos in Video-MME with 16 threads.}
  \label{fig:duration_lat}
\end{figure}

\begin{table*}[t]
  \centering
  \caption{Accuracy (\%) and preprocessing latency on Video-MME, MLVU, and LongVideoBench (LVB). Preproc.\ is the per-video latency of RAPID-Dec decoding and CLIP ViT-B/32 scoring, averaged over three runs on 10\% of the videos of each benchmark. 
  }
  \vspace{-8pt}
  \small
  \setlength{\tabcolsep}{5pt}
  \begin{tabular}{l c c cccc cccc}
    \toprule
    & & & \multicolumn{4}{c}{Qwen3-VL-8B} & \multicolumn{4}{c}{GLM-4.6V-Flash} \\
    \cmidrule(lr){4-7} \cmidrule(lr){8-11}
    Method & \#Frames & Preproc. (s) $\downarrow$
      & Video-MME & MLVU & LVB & Avg.
      & Video-MME & MLVU & LVB & Avg. \\
    \midrule
    Uniform   & 32 & 0.60 & 60.26 & 60.69 & 55.95 & 58.97 & 57.15 & 58.68 & 55.27 & 57.03 \\
    Top-K     & 32 & 3.41 & 61.56 & 68.81 & 57.89 & 62.75 & 59.70 & 67.08 & 60.13 & 62.30 \\
    AKS       & 32 & 3.41 & 61.44 & 65.84 & 56.99 & 61.42 & 59.26 & 61.70 & 57.67 & 59.54 \\
    Focus     & 32 & 2.36 & 61.04 & 67.40 & 58.27 & 62.24 & 58.30 & 64.76 & 60.21 & 61.09 \\
    WFS-SB    & 32 & 3.41 & 62.78 & 66.81 & 58.26 & 62.62 & 60.93 & 64.33 & 58.49 & 61.25 \\
    RAPID-Sel & 32 & 1.12 & 61.48 & 67.71 & 57.67 & 62.29 & 60.30 & 65.65 & 59.54 & 61.83 \\
    \midrule
    Uniform   & 64 & 0.83 & 63.63 & 63.72 & 58.86 & 62.07 & 59.96 & 62.14 & 58.34 & 60.15 \\
    Top-K     & 64 & 3.41 & 63.81 & 70.50 & 59.91 & 64.74 & 62.63 & 68.05 & 61.78 & 64.15 \\
    AKS       & 64 & 3.41 & 65.70 & 69.93 & 60.36 & 65.33 & 62.41 & 66.19 & 60.81 & 63.14 \\
    Focus     & 64 & 2.39 & 64.26 & 70.07 & 60.73 & 65.02 & 61.04 & 66.58 & 62.08 & 63.23 \\
    WFS-SB    & 64 & 3.41 & 65.26 & 69.89 & 59.91 & 65.02 & 62.93 & 66.22 & 60.36 & 63.17 \\
    RAPID-Sel & 64 & 1.30 & 65.41 & 70.05 & 60.58 & 65.35 & 62.44 & 66.71 & 61.71 & 63.62 \\
    \bottomrule
  \end{tabular}
  \label{tab:main}
\end{table*}

\medskip
\noindent\textbf{Candidate sampling across video durations.}
Many frame selection methods decode candidate frames at 1~fps, so the number of candidates grows linearly with video duration.
Fig.~\ref{fig:duration_lat} shows their decoding latency for H.264/AVC videos of 5 to 60 minutes.
RAPID-Dec achieves the lowest latency for every duration, with 3.6--4.3$\times$ and 1.8--2.2$\times$ speedups over TorchCodec and QuickCodec, respectively.
For hour-long videos, RAPID-Dec takes 9.8~s, whereas TorchCodec and Decord take 40.5~s and 45.1~s, respectively.
Since RAPID-Dec is agnostic to the choice of target frames, existing frame selection methods can directly benefit from it.

\subsection{Accuracy and efficiency of RAPID-Sel}\label{sec:exp-sel}
We next compare RAPID-Sel with existing frame selection methods at the same frame budget, using RAPID-Dec for all methods to isolate frame selection from the decoder choice.

\medskip
\noindent\textbf{Main results.}
As shown in Table~\ref{tab:main}, all frame selection methods improve the average accuracy over uniform sampling, which confirms the benefit of query-aware frame selection for long videos.
RAPID-Sel gains 3.32 and 3.28 points with Qwen3-VL-8B, and 4.80 and 3.47 points with GLM-4.6V-Flash, using 32 and 64 frames, respectively.
RAPID-Sel achieves the highest average accuracy with Qwen3-VL-8B and 64 frames, and stays within 0.53 points of the best method in the other three settings.
Meanwhile, RAPID-Sel takes only 1.12~s and 1.30~s for preprocessing with 32 and 64 frames, which is 3.0$\times$ and 2.6$\times$ faster than Top-K, AKS, and WFS-SB, and 2.1$\times$ and 1.8$\times$ faster than Focus, respectively.
This is only about 0.5~s more than uniform sampling.
Since all methods use RAPID-Dec, this speedup comes from decoding and scoring only the keyframes instead of dense candidates.

\begin{table}[b]
  \centering
  \caption{Uniform vs.\ keyframe candidates of RAPID-Sel, equal in number, on Video-MME with Qwen3-VL-8B and 64 frames.}
  \vspace{-8pt}
  \small
  \setlength{\tabcolsep}{3pt}
  \begin{tabular}{l cccc}
    \toprule
    Candidates & Short & Medium & Long & Avg. \\
    \midrule
    Uniform & \textbf{76.22} & 63.67 & 55.22 & 65.04 \\
    Keyframes     & 75.67 & \textbf{63.78} & \textbf{56.78} & \textbf{65.41} \\
    \bottomrule
  \end{tabular}
  \label{tab:cand-uniform}
\end{table}

\medskip
\noindent\textbf{Impact of keyframe candidates.}
To evaluate the benefit of keyframe-based selection, we replace the keyframes of each video with the same number of evenly spaced frames, treat the frames from each candidate to the next as its group, and keep the rest of RAPID-Sel unchanged. As shown in Table~\ref{tab:cand-uniform}, keyframe candidates achieve slightly higher average accuracy (65.41 vs.\ 65.04), with the difference increasing from $-0.55$ points on short videos to $+1.56$ points on long videos. This suggests that keyframe-based candidates are particularly beneficial for long videos, where sparse uniform candidates may miss relevant visual changes.

\begin{figure}[t]
  \centering
  \begin{subfigure}{0.30\linewidth}
    \includegraphics[width=\linewidth]{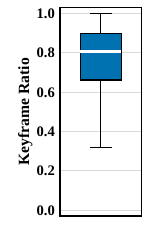}
    \caption{Placement}
    \label{fig:scene-key-a}
  \end{subfigure}
  \hfill
  \begin{subfigure}{0.68\linewidth}
    \includegraphics[width=\linewidth]{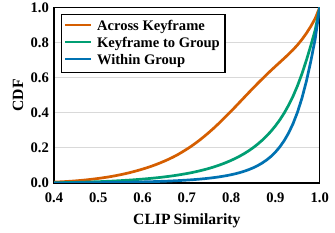}
    \caption{Similarity}
    \label{fig:scene-key-b}
  \end{subfigure}
  \vspace{-8pt}
  \caption{
  Keyframe placement and visual consistency over 2,098 videos of Video-MME, LongVideoBench, and MLVU. (a) Per-video fraction of low-similarity 1~s intervals (CLIP similarity below 0.73) containing a keyframe. (b) CDFs of CLIP similarity between adjacent 1~fps frames within a group, across a keyframe, and between each keyframe and the other frames of its group.
  }
  \label{fig:scene-key}
\end{figure}
\medskip
\noindent\textbf{Keyframes and visual consistency.}
We examine how keyframes relate to visual changes using the CLIP ViT-B/32 embeddings of the 1~fps frames in the three benchmarks.
For each video, we compute the cosine similarity between every pair of its 1~fps frames and take the median, which indicates how similar two arbitrary moments of the video are. 
The median of this value over all videos is 0.73, and we therefore identify the 1~s interval between two adjacent 1~fps frames as a low-similarity interval if their similarity is below 0.73 (7.8\% of all intervals).
As shown in Fig.~\ref{fig:scene-key-a}, the median per-video fraction of these intervals containing a keyframe is 0.81, indicating that keyframes are placed at most substantial visual changes.

In Fig.~\ref{fig:scene-key-b}, a similarity below 0.73 is 11$\times$ more frequent between adjacent 1~fps frames across a keyframe than within a group (24.7\% vs.\ 2.2\%), whereas 82.4\% of the pairs within a group exceed 0.9.
Some pairs across a keyframe are still highly similar, likely because encoders also insert keyframes when a group reaches its maximum length~\citep{x264,x265}.
To check whether a keyframe represents its group, we also compare it with the other 1~fps frames of its group.
These similarities resemble those within a group rather than those across a keyframe, and are only slightly lower, likely because they span longer temporal intervals over which gradual visual changes can accumulate.
These results support the use of keyframes as coarse representatives of their groups.

\medskip
\noindent\textbf{Impact of $\rho$.}
\begin{table}[t]
  \centering
  \caption{Impact of $\rho$ (Video-MME, Qwen3-VL-8B, 64 frames).
  }
  \vspace{-8pt}
  \small
  \setlength{\tabcolsep}{7pt}
  \begin{tabular}{lcccc}
    \toprule
    $\rho$ & Short & Medium & Long & Avg. \\
    \midrule
    0.45 & 75.44 & 63.67 & 55.22 & 64.78 \\
    0.50 & 75.67 & \textbf{63.78} & \textbf{56.78} & \textbf{65.41} \\
    0.55 & 75.22 & 63.22 & 56.11 & 64.85 \\
    0.60 & \textbf{76.00} & 61.89 & 55.22 & 64.37 \\
    \midrule
    Uni. ($\rho=1$) & 74.56 & 61.89 & 54.44 & 63.63 \\
    \bottomrule
  \end{tabular}
  \label{tab:rho}
\end{table}
Table~\ref{tab:rho} shows the accuracy on Video-MME for different values of the uniform ratio $\rho$. Every tested value outperforms uniform sampling ($\rho=1$) on average, showing that allocating part of the frame budget to query-relevant groups is beneficial. The average accuracy peaks at our default $\rho=0.50$, 1.78 points above uniform sampling, mainly due to the gains on medium and long videos. As $\rho$ decreases from 0.60 to 0.50, the relevance budget grows and the accuracy on long videos improves from 55.22 to 56.78. Decreasing $\rho$ to 0.45 lowers the accuracy, likely because fewer uniform frames remain to cover the entire video.

\subsection{End-to-end efficiency of RAPID}
\label{sec:exp-e2e}
\begin{figure}[t]
  \centering
  \includegraphics[width=\linewidth]{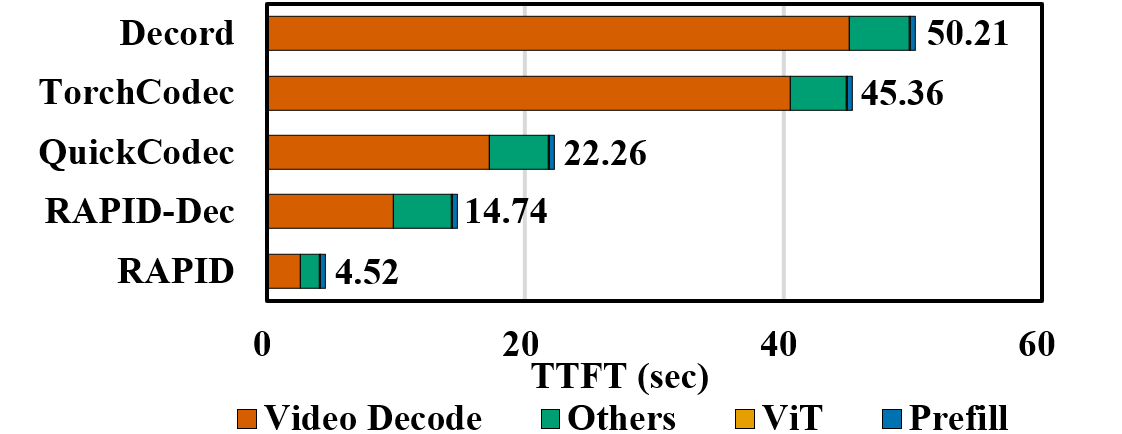}
  \vspace{-18pt}
  \caption{TTFT breakdown of Qwen3-VL-8B with 64 frames on hour-long videos. The first four bars use AKS with each decoder. Others includes relevance scoring and frame selection.}
  \label{fig:e2e_lat}
\end{figure}
Fig.~\ref{fig:e2e_lat} shows the TTFT breakdown of Qwen3-VL-8B with 64 frames on the hour-long videos in Sec.~\ref{sec:exp-dec},
comparing RAPID with AKS, the most accurate prior method with Qwen3-VL-8B and 64 frames (Table~\ref{tab:main}).
With AKS, video decoding dominates the TTFT, taking 40.5~s of 45.4~s with TorchCodec.
RAPID-Dec reduces it to 9.8~s, achieving 3.1$\times$ and 1.5$\times$ TTFT speedups over TorchCodec and QuickCodec, respectively, but decoding and scoring all candidates at 1~fps still take 14.3~s.
RAPID-Sel reduces this cost to 4.1~s by decoding and scoring only the keyframes.
As a result, RAPID reduces the TTFT to 4.5~s, achieving 3.3$\times$, 4.9$\times$, and 10.0$\times$ speedups over AKS with RAPID-Dec, QuickCodec, and TorchCodec, respectively, with accuracy comparable to AKS (65.35\% vs.\ 65.33\%).

\section{Conclusion}
\label{sec:conclusion}
We presented RAPID, a codec-aware video decoding framework for efficient long-video understanding.
We showed that video decoding is a major bottleneck of TTFT, especially for frame selection methods that decode dense candidate frames.
RAPID exploits the bitstream through two components.
For any targets, RAPID-Dec skips unnecessary frames using keyframe and reference information and decodes the rest in parallel without approximation, up to 3.7$\times$ faster than the fastest baseline on H.264/AVC.
RAPID-Sel scores keyframes instead of dense candidates to select query-relevant frames, with comparable accuracy and up to a 3.0$\times$ preprocessing speedup.
Combining both components, RAPID achieves a 4.9$\times$ speedup in TTFT on hour-long videos over a state-of-the-art frame selection method with the fastest baseline decoder.

\section*{LLM Usage Statement}
We used Claude (Anthropic) in preparing this manuscript. The authors drafted the manuscript, and Claude was used to refine grammar, wording, and sentence structure, including the related work and experimental results sections. We also provided full manuscript drafts to Claude for reviewer-style feedback, and the authors independently evaluated and decided which suggestions to incorporate.
Claude assisted in searching for relevant prior work, while the authors verified all cited references against the original publications and made the final citation decisions. Claude was also used to generate portions of the implementation based on author-provided specifications. All generated code was reviewed, tested, and modified as necessary by the authors.
The authors independently designed and conducted all experiments and analyzed the results. Claude did not contribute to the development of research ideas, hypotheses, substantive claims, or conclusions, nor was it used to create figures, diagrams, or other visual materials. No LLM tools were used for synthetic data generation, dataset creation, or annotation.
The authors take full responsibility for all content of this paper.
{
    \small
    \bibliographystyle{ieeenat_fullname}
    \bibliography{arxiv}

@String(CVPR= {IEEE Conf. Comput. Vis. Pattern Recog.})

@String(ICCV= {Int. Conf. Comput. Vis.})

@String(NIPS= {Adv. Neural Inform. Process. Syst.})

@String(ICLR = {Int. Conf. Learn. Represent.})

@String(TCSVT = {IEEE Trans. Circuit Syst. Video Technol.})

@String(ICML  = {Int. Conf. Mach. Learn.})

@String(TMLR  = {Trans. Mach. Learn. Res.})

@String(CVPR  = {IEEE Conf. Comput. Vis. Pattern Recog. (CVPR)})

@String(ICCV  = {Int. Conf. Comput. Vis. (ICCV)})

@String(NIPS  = {Adv. Neural Inform. Process. Syst. (NeurIPS)})

@String(ICLR  = {Int. Conf. Learn. Represent. (ICLR)})

@String(ICML  = {Int. Conf. Mach. Learn. (ICML)})

@String(TMLR  = {Trans. Mach. Learn. Res. (TMLR)})

@String(TCSVT = {IEEE Trans. Circuits Syst. Video Technol. (TCSVT)})

@article{Qwen3-VL,
  title={{Qwen3-VL} technical report},
  author={Bai, Shuai and Cai, Yuxuan and Chen, Ruizhe and Chen, Keqin and Chen, Xionghui and Cheng, Zesen and Deng, Lianghao and Ding, Wei and Gao, Chang and Ge, Chunjiang and others},
  journal={arXiv preprint arXiv:2511.21631},
  year={2025}
}

@article{Qwen2_5_VL,
  title={{Qwen2.5-VL} Technical Report},
  author={Bai, Shuai and Chen, Keqin and Liu, Xuejing and Wang, Jialin and Ge, Wenbin and Song, Sibo and Dang, Kai and Wang, Peng and Wang, Shijie and Tang, Jun and others},
  journal={arXiv preprint arXiv:2502.13923},
  year={2025}
}

@article{glm46,
  title={{GLM-4.5V} and {GLM-4.1V-Thinking}: Towards versatile multimodal reasoning with scalable reinforcement learning},
  author={Hong, Wenyi and Yu, Wenmeng and Gu, Xiaotao and Wang, Guo and Gan, Guobing and Tang, Haomiao and Cheng, Jiale and Qi, Ji and Ji, Junhui and Pan, Lihang and others},
  journal={arXiv preprint arXiv:2507.01006},
  year={2025}
}

@article{wang2025internvl3_5,
  title={{InternVL3.5}: Advancing Open-Source Multimodal Models in Versatility, Reasoning, and Efficiency},
  author={Wang, Weiyun and Gao, Zhangwei and Gu, Lixin and Pu, Hengjun and Cui, Long and Wei, Xingguang and Liu, Zhaoyang and Jing, Linglin and Ye, Shenglong and Shao, Jie and others},
  journal={arXiv preprint arXiv:2508.18265},
  year={2025}
}

@inproceedings{clip,
  title={Learning Transferable Visual Models from Natural Language Supervision},
  author={Radford, Alec and Kim, Jong Wook and Hallacy, Chris and Ramesh, Aditya and Goh, Gabriel and Agarwal, Sandhini and Sastry, Girish and Askell, Amanda and Mishkin, Pamela and Clark, Jack and others},
  booktitle=ICML,
  year={2021}
}

@inproceedings{li2022blip,
      title={{BLIP}: Bootstrapping Language-Image Pre-training for Unified Vision-Language Understanding and Generation}, 
      author={Junnan Li and Dongxu Li and Caiming Xiong and Steven Hoi},
      booktitle=ICML,
      year={2022}
}

@article{
schneider2026quickvideo,
title={{QuickVideo}: Real-Time Long Video Understanding with System Algorithm Co-Design},
author={Benjamin Schneider and Dongfu Jiang and Chao Du and Tianyu Pang and Wenhu Chen},
journal=TMLR,
year={2026}
}

@article{zhao2025flashcodec,
  title={Enabling Disaggregated Multi-Stage {MLLM} Inference via {GPU}-Internal Scheduling and Resource Sharing},
  author={Zhao, Lingxiao and Zhou, Haoran and Che, Yuezhi and Cheng, Dazhao},
  journal={arXiv preprint arXiv:2512.17574},
  year={2025}
}

@misc{decord2019,
  author       = {{Distributed (Deep) Machine Learning Community}},
  title        = {Decord: An Efficient Video Loader for Deep Learning},
  howpublished = {\url{https://github.com/dmlc/decord}},
  year         = {2019},
  note         = {Accessed: 2026-11-13}
}

@misc{torchcodec,
  author       = {{TorchCodec Contributors}},
  title        = {{TorchCodec}: Media Decoding and Encoding for {PyTorch}},
  howpublished = {\url{https://github.com/meta-pytorch/torchcodec}},
  year         = {2024},
  note         = {Version 0.15.0. Accessed: 2026-11-13}
}

@inproceedings{bolya2022tome,
  title={Token Merging: Your {ViT} but Faster},
  author={Bolya, Daniel and Fu, Cheng-Yang and Dai, Xiaoliang and Zhang, Peizhao and Feichtenhofer, Christoph and Hoffman, Judy},
  booktitle=ICLR,
  year={2023}
}

@inproceedings{yang2025visionzip,
  title={{VisionZip}: Longer is Better but Not Necessary in Vision Language Models},
  author={Yang, Senqiao and Chen, Yukang and Tian, Zhuotao and Wang, Chengyao and Li, Jingyao and Yu, Bei and Jia, Jiaya},
  booktitle=CVPR,
  year={2025}
}

@inproceedings{
tang2026surge,
title={{SURGE}: Surprise-Guided Token Reduction for Efficient Video Understanding with {VLM}s},
author={Chong Tang and Sannara Ek and Dirk Koch and Robert D. Mullins and Alex S. Weddell and Jagmohan Chauhan},
booktitle=ICLR,
year={2026}
}

@article{zou2026codecsight,
  title={{CodecSight}: Leveraging Video Codec Signals for Efficient Streaming {VLM} Inference},
  author={Zou, Yulin and Chen, Wenyan and Chen, Yan and Rajan, Anya and Park, JooYoung and Nitin, Shivaraman and Tao, Luo and Romero, Francisco and Ustiugov, Dmitrii},
  journal={arXiv preprint arXiv:2604.06036},
  year={2026}
}

@inproceedings{
shao2025holitom,
title={{HoliTom}: Holistic Token Merging for Fast Video Large Language Models},
author={Kele Shao and Keda Tao and Can Qin and Haoxuan You and Yang Sui and Huan Wang},
booktitle=NIPS,
year={2025}
}

@inproceedings{
    wang2026earlytom,
    title={{EarlyTom}: Early Token Compression Completes Fast Video Understanding},
    author={Wang, Hesong and Jin, Xin and Lu, Lu and Li, Chenhaowen and Chen, Jian and Liu, Qiang and Wang, Huan},
    booktitle=CVPR,
    year={2026}
}

@inproceedings{tao2025dycoke,
  title={{DyCoke}: Dynamic compression of tokens for fast video large language models},
  author={Tao, Keda and Qin, Can and You, Haoxuan and Sui, Yang and Wang, Huan},
  booktitle=CVPR,
  year={2025}
}

@inproceedings{mllm_select_hu2025m,
  title={{M-LLM} based video frame selection for efficient video understanding},
  author={Hu, Kai and Gao, Feng and Nie, Xiaohan and Zhou, Peng and Tran, Son and Neiman, Tal and Wang, Lingyun and Shah, Mubarak and Hamid, Raffay and Yin, Bing and others},
  booktitle=CVPR,
  year={2025}
}

@inproceedings{
rl_frame_select_qin2026efficient,
title={Efficient Frame Selection for Long Video Understanding via Reinforcement Learning},
author={Yaxuan Qin and Hefei Li and Wenqi Mu and Yancheng He},
booktitle=CVPR,
year={2026},
}

@inproceedings{voyager_yu2025frame,
  title={{Frame-Voyager}: Learning to query frames for video large language models},
  author={Yu, Sicheng and Jin, Chengkai and Wang, Huanyu and Chen, Zhenghao and Jin, Sheng and Zuo, Zhongrong and Xu, Xiaolei and Sun, Zhenbang and Zhang, Bingni and Wu, Jiawei and others},
  booktitle=ICLR,
  year={2025}
}

@inproceedings{aks,
  title={Adaptive Keyframe Sampling for Long Video Understanding},
  author={Tang, Xi and Qiu, Jihao and Xie, Lingxi and Tian, Yunjie and Jiao, Jianbin and Ye, Qixiang},
  booktitle=CVPR,
  year={2025}
}

@inproceedings{
ziruiz2026focus,
title={{FOCUS}: Efficient Keyframe Selection for Long Video Understanding},
author={Zirui Zhu and Hailun Xu and Yang Luo and Yong Liu and Kanchan Sarkar and Zhenheng Yang and Yang You},
booktitle=ICLR,
year={2026}
}

@inproceedings{
wfs_bs,
title={Wavelet-based Frame Selection by Detecting Semantic Boundary for Long Video Understanding},
author={Wang Chen and Yuhui Zeng and Yongdong Luo and Tianyu Xie and Luojun Lin and Jiayi Ji and Yan Zhang and Xiawu Zheng},
booktitle=CVPR,
year={2026}
}

@inproceedings{zhou2025mlvu,
  title={{MLVU}: Benchmarking Multi-Task Long Video Understanding},
  author={Zhou, Junjie and Shu, Yan and Zhao, Bo and Wu, Boya and Liang, Zhengyang and Xiao, Shitao and Qin, Minghao and Yang, Xi and Xiong, Yongping and Zhang, Bo and others},
  booktitle=CVPR,
  year={2025}
}

@inproceedings{wu2024longvideobench,
  title={{LongVideoBench}: A Benchmark for Long-Context Interleaved Video-Language Understanding},
  author={Wu, Haoning and Li, Dongxu and Chen, Bei and Li, Junnan},
  booktitle=NIPS,
  year={2024}
}

@inproceedings{videomme,
  title={{Video-MME}: The First-Ever Comprehensive Evaluation Benchmark of Multi-Modal {LLMs} in Video Analysis},
  author={Fu, Chaoyou and Dai, Yuhan and Luo, Yongdong and Li, Lei and Ren, Shuhuai and Zhang, Renrui and Wang, Zihan and Zhou, Chenyu and Shen, Yunhang and Zhang, Mengdan and others},
  booktitle=CVPR,
  year={2025}
}

@article{avc_overview,
  title={Overview of the {H.264/AVC} Video Coding Standard}, 
  author={Wiegand, T. and Sullivan, G.J. and Bjontegaard, G. and Luthra, A.},
  journal = TCSVT,
  volume  = {13},
  number  = {7},
  pages   = {560--576},
  year    = {2003}
  }

@article{hevc_overview,
  title   = {Overview of {HEVC} High-Level Syntax and Reference Picture Management},
  author={Sj{\"o}berg, Rickard and Chen, Ying and Fujibayashi, Akira and Hannuksela, Miska M. and Samuelsson, Jonatan and Tan, Thiow Keng and Wang, Ye-Kui and Wenger, Stephan},
  journal = TCSVT,
  volume  = {22},
  number  = {12},
  pages   = {1858--1870},
  year    = {2012}
  }

@article{wang2021vvc_hls,
  title   = {The High-Level Syntax of the Versatile Video Coding ({VVC}) Standard},
  author  = {Wang, Ye-Kui and Skupin, Robert and Hannuksela, Miska M. and Deshpande, Sachin and Hendry and Drugeon, Virginie and Sj{\"o}berg, Rickard and Choi, Byeongdoo and Seregin, Vadim and Sanchez, Yago and Boyce, Jill M. and Wan, Wade and Sullivan, Gary J.},
  journal = TCSVT,
  volume  = {31}, number = {10}, pages = {3779--3800}, year = {2021}
}

@article{av1,
title   = {A Technical Overview of {AV1}},
author = {Han, Jingning and Li, Bohan and Mukherjee, Debargha and Chiang, Ching-Han and Grange, Adrian and Chen, Cheng and Su, Hui and Parker, Sarah and Deng, Sai and Joshi, Urvang and Chen, Yue and Wang, Yunqing and Wilkins, Paul and Xu, Yaowu and Bankoski, James},
journal = {Proc. IEEE},
volume  = {109},
number  = {9},
pages   = {1435--1462},
year    = {2021}
}

@article{adpativeiframe,
  title   = {Adaptive Intra-Frame Assignment and Bit-Rate Estimation for Variable {GOP} Length in {H.264}},
  author={Lee, Jeehong and Shin, Ilhong and Park, Hyunwook},
  journal=TCSVT, 
  volume  = {16},
  number  = {10},
  pages   = {1271--1279},
  year    = {2006}}

@inproceedings{sastre2006shotiframe,
  title     = {Shot Detection Method for Low Bit-Rate {H.264} Video Coding},
  author    = {Sastre, J. and Usach, P. and Moya, A. and Naranjo, V. and Lopez, J. M.},
  booktitle = {Eur. Signal Process. Conf. (EUSIPCO)},
  year      = {2006}
}

@misc{x264,
  author       = {{VideoLAN}},
  title        = {x264: A free {H.264/AVC} encoder},
  howpublished = {\url{https://www.videolan.org/developers/x264.html}},
  note         = {Accessed: 2026-11-15}
}

@misc{x265,
  author       = {{MulticoreWare}},
  title        = {x265 Documentation: Command Line Options},
  howpublished = {\url{https://x265.readthedocs.io/en/4.1/cli.html}},
  note         = {Version 4.1. Accessed: 2026-11-15}
}

@inproceedings{mukherjee2013vp9,
  title     = {The Latest Open-Source Video Codec {VP9} -- An Overview and Preliminary Results},
  author    = {Mukherjee, Debargha and Bankoski, Jim and Grange, Adrian and Han, Jingning and Koleszar, John and Wilkins, Paul and Xu, Yaowu and Bultje, Ronald},
  booktitle = {Picture Coding Symp. (PCS)},
  year      = {2013}
}

@article{balinski1978jefferson_dhondt,
  title   = {The {J}efferson Method of Apportionment},
  author  = {Balinski, M. L. and Young, H. P.},
  journal = {SIAM Review},
  volume  = {20},
  number  = {2},
  pages   = {278--284},
  year    = {1978}
}

@inproceedings{lmms_eval2024,
    title = "{LMMs-Eval}: Reality Check on the Evaluation of Large Multimodal Models",
    author = "Zhang, Kaichen  and
      Li, Bo  and
      Zhang, Peiyuan  and
      Pu, Fanyi  and
      Cahyono, Joshua Adrian  and
      Hu, Kairui  and
      Liu, Shuai  and
      Zhang, Yuanhan  and
      Yang, Jingkang  and
      Li, Chunyuan  and
      Liu, Ziwei",
    editor = "Chiruzzo, Luis  and
      Ritter, Alan  and
      Wang, Lu",
    booktitle = {Findings Assoc. Comput. Linguist.: NAACL},
    year = {2025}
}

@article{sarkar2026copevideolm,
  title={{CoPE-VideoLM}: Leveraging Codec Primitives for Efficient Video Language Modeling},
  author={Sarkar, Sayan Deb and Pautrat, R{\'e}mi and Miksik, Ondrej and Pollefeys, Marc and Armeni, Iro and Rad, Mahdi and Dusmanu, Mihai},
  journal={arXiv preprint arXiv:2602.13191},
  year={2026}
}

@inproceedings{
yashima2026remora,
title={{ReMoRa}: Multimodal Large Language Model based on Refined Motion Representation for Long-Video Understanding},
author={Daichi Yashima and Shuhei Kurita and Yusuke Oda and Komei Sugiura},
booktitle=CVPR,
year={2026}
}

@inproceedings{beforefirsttoken,
title = {Before the First Token: Benchmarking Data Preprocessing in Vision-Language Models},
author = {Zohdi, Sepideh and Wang, Lin},
booktitle = {Eur. Workshop Mach. Learn. Syst. (EuroMLSys)},
year = {2026}
}

@article{smol,
title   = {Jointly Optimizing Preprocessing and Inference for {DNN}-Based Visual Analytics},
author = {Kang, Daniel and Mathur, Ankit and Veeramacheneni, Teja and Bailis, Peter and Zaharia, Matei},
journal = {Proc. VLDB Endow. (PVLDB)},
volume = {14},
number = {2},
pages = {87–100},
year = {2020}
}

@inproceedings{hwang2022cova,
  title     = {{CoVA}: Exploiting Compressed-Domain Analysis to Accelerate Video Analytics},
  author    = {Hwang, Jinwoo and Kim, Minsu and Kim, Daeun and Nam, Seungho and Kim, Yoonsung and Kim, Dohee and Sharma, Hardik and Park, Jongse},
  booktitle = {USENIX Annu. Tech. Conf. (ATC)},
  year      = {2022}
}

@inproceedings{coviar,
  title={Compressed Video Action Recognition},
  author={Wu, Chao-Yuan and Zaheer, Manzil and Hu, Hexiang and Manmatha, R and Smola, Alexander J and Kr{\"a}henb{\"u}hl, Philipp},
  booktitle=CVPR,
  year={2018}
}

@inproceedings{korbar2019scsampler,
  title={{SCSampler}: Sampling Salient Clips from Video for Efficient Action Recognition},
  author={Korbar, Bruno and Tran, Du and Torresani, Lorenzo},
  booktitle=ICCV,
  year={2019}
}

@misc{ffmpeg,
  author       = {{FFmpeg Developers}},
  title        = {{FFmpeg} Codecs Documentation: Codec Options},
  howpublished = {\url{https://ffmpeg.org/ffmpeg-codecs.html}},
  year         = {2026},
  note         = {Accessed: 2026-11-15}
}
}

\clearpage
\maketitlesupplementary
\appendix
\noindent
In the supplementary material, we provide additional implementation details, experiments, analyses, and visualizations. The supplementary material is organized as follows:

\begin{itemize}
\item 
Sec.~\ref{sec:appx-impl} describes the implementation details and how RAPID-Dec identifies keyframes and non-reference frames in each codec.
\item
Sec.~\ref{sec:appx-exp} evaluates RAPID-Dec on different hardware and on all Video-MME videos, reports the candidate overhead of RAPID-Sel, and compares the accuracy--TTFT trade-off of RAPID with those of existing methods.
\item 
Sec.~\ref{sec:appx-analysis} provides additional analysis on how keyframes relate to visual changes, with visualized examples.
\item
Sec.~\ref{sec:appx-future} discusses the limitations and future directions of RAPID.
\end{itemize}

\section{Implementation details}\label{sec:appx-impl}
\subsection{Detailed evaluation settings}
\noindent\textbf{Frame selection methods.}
Top-K, AKS~\citep{aks}, and WFS-SB~\citep{wfs_bs} score candidate frames sampled at 1~fps over the entire video. Top-K selects the frames with the highest scores, while AKS and WFS-SB select frames with the algorithms of their official implementations and default hyperparameters. We run the official implementation of Focus~\citep{ziruiz2026focus} to identify query-relevant regions. Within these regions, top-scoring frames are selected. For RAPID-Sel, $\tau$ in Eq.~\eqref{eq:logit} is the logit scale of the pretrained CLIP model, $\tau=100$. If a short video has fewer 1~fps frames than the frame budget, all frame selection methods, including RAPID-Sel, fall back to uniform sampling.

\noindent\textbf{Model input.}
Before being fed to the VLMs, each frame is resized with its aspect ratio preserved to at most 65,536 pixels, so that both models receive nearly the same number of pixels.
Qwen3-VL-8B~\citep{Qwen3-VL} maps each $32\times32$ region of two consecutive frames to one visual token, so each side is rounded to a multiple of 32. GLM-4.6V-Flash~\citep{glm46} uses $28\times28$ regions instead, so each side is rounded to the nearest multiple of 28.

\noindent\textbf{Videos for Figs.~\ref{fig:parallel_lat} and~\ref{fig:sample_lat}.}
The four hour-long videos are taken from the 10\% longest videos of Video-MME~\citep{videomme}. To cover different group lengths, we choose two videos with long keyframe groups (149 and 158 frames on average) and two with short ones (47 and 63 frames). We re-encode them into H.265/HEVC and VP9 with FFmpeg~\citep{ffmpeg}, using libx265 and libvpx-vp9, respectively. Each re-encoded video keeps the resolution, frame rate, and pixel format of the original and targets the bitrate of the original video stream. 
We force an intra frame at each IDR frame of the original, prevent the encoders from inserting additional keyframes at scene changes, and keep the default open group of pictures (GOP) setting of x265, which allows frames that follow an intra frame in decoding order to reference frames preceding it.
For VP9, we disable alternate reference frames, since we observed that Decord fails to decode them with multiple threads.

\subsection{RAPID-Dec on various codecs} \label{sec:codec-impl}
As noted in Sec.~\ref{sec:codec-explain}, each codec defines its own bitstream syntax, including the syntax elements that indicate keyframes and reference frames.
We describe how RAPID-Dec identifies keyframes and non-reference frames in each codec.
We begin with H.264/AVC, in which 98.0\% of the videos in Video-MME~\citep{videomme}, LongVideoBench~\citep{wu2024longvideobench}, and MLVU~\citep{zhou2025mlvu} are encoded, followed by H.265/HEVC, in which most of the remaining videos are encoded.

\noindent\textbf{H.264/AVC.}
Each NAL unit of H.264/AVC~\citep{avc_overview} starts with a one-byte header (Fig.~\ref{fig:codec-nal}) containing nal\_unit\_type, which indicates the type of data the NAL unit carries, and nal\_ref\_idc. A nal\_unit\_type of 5 indicates a slice of an IDR frame, and a nal\_ref\_idc of 0 in the header of a slice indicates a non-reference frame. Besides IDR frames, an encoder can mark an I-frame as a random access point with a recovery point message carried in a supplemental enhancement information (SEI) NAL unit (nal\_unit\_type 6), referred to as a recovery point SEI. RAPID-Dec uses IDR frames as keyframes, as well as I-frames whose recovery point SEI guarantees that decoding from them yields the same frames as decoding the entire video, namely those with recovery\_frame\_cnt $=0$, exact\_match\_flag $=1$, and broken\_link\_flag $=0$.

In an H.264/AVC open GOP, the B-frames that follow an I-frame with a recovery point SEI in decoding order but precede it in display order may reference the preceding group.
Our implementation does not support such streams. However, open GOPs are disabled by default in x264~\citep{x264}, and none of the benchmark videos contains such frames.

\noindent\textbf{H.265/HEVC.}
Like H.264/AVC, an H.265/HEVC bitstream~\citep{hevc_overview} consists of NAL units, but each starts with a two-byte header that contains nal\_unit\_type and TemporalId. H.265/HEVC defines three types of intra random access point (IRAP) frames: IDR, clean random access (CRA), and broken link access (BLA). 
Random access skipped leading (RASL) frames may follow a CRA or BLA frame in decoding order and may reference frames of the previous group.
Since decoding from such an IRAP frame cannot reconstruct its RASL frames, RAPID-Dec uses an IRAP frame as a keyframe only if no RASL frame appears between it and the next IRAP frame in decoding order. For non-IRAP frames, nal\_unit\_type also carries the reference information, since each type has two variants whose names end with \_N and \_R.
An \_N frame is not referenced by subsequent frames of the same temporal sub-layer, but may be referenced by frames of higher sub-layers. RAPID-Dec therefore treats an \_N frame as a non-reference frame only if its TemporalId is the highest in the video.

\noindent\textbf{VP9.}
VP9~\citep{mukherjee2013vp9} is an open-source video codec developed outside the standardization of H.264/AVC and H.265/HEVC, and we include it to examine whether RAPID-Dec generalizes beyond NAL-based codecs. Instead of NAL units, each VP9 frame starts with an uncompressed header, in which a single bit indicates whether the frame is a \texttt{KEY\_FRAME}, the VP9 frame type that resets the decoding state. Since such a frame satisfies our definition of a keyframe, RAPID-Dec uses every \texttt{KEY\_FRAME} as a keyframe, which it identifies from the keyframe flag that the container stores for each packet. Whether a frame can be referenced is given by refresh\_frame\_flags in the header. A frame with refresh\_frame\_flags $=0$ is not stored as a reference frame. 
However, unlike a non-reference frame in H.264/AVC and H.265/HEVC, such a frame can still affect the decoding of subsequent frames, since the next frame may use its motion vectors or segmentation map.
As refresh\_frame\_flags alone thus cannot determine whether a frame can be skipped, RAPID-Dec treats every VP9 frame as a reference frame, so only (i) and (ii) in Sec.~\ref{sec:rapid-dec} apply.

\section{Extended evaluations}\label{sec:appx-exp}
\subsection{RAPID-Dec on different hardware}
To verify that the efficiency of RAPID-Dec does not depend on a specific hardware setup, we evaluate it on NVDEC and on another CPU, using the same videos as in Fig.~\ref{fig:parallel_lat} and Fig.~\ref{fig:sample_lat}.

\begin{figure*}[t]
  \centering
  \begin{subfigure}{0.32\linewidth}
    \includegraphics[width=\linewidth]{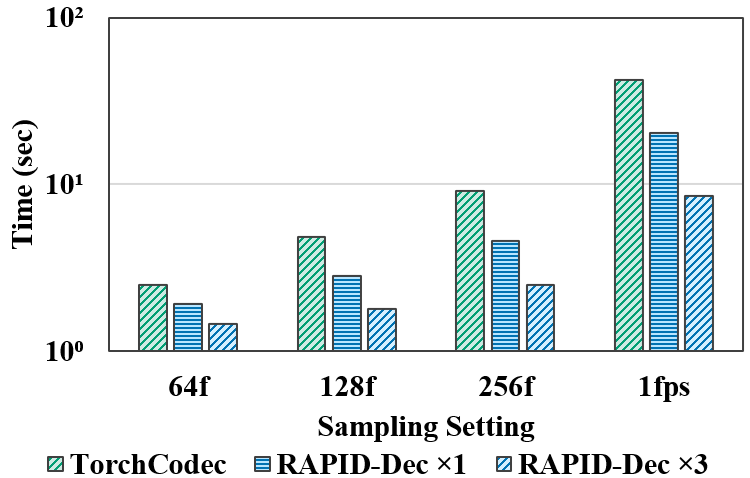}
    \caption{H.264/AVC}
    \label{fig:nvdec_lat-a}
  \end{subfigure}
  \hfill
  \begin{subfigure}{0.32\linewidth}
    \includegraphics[width=\linewidth]{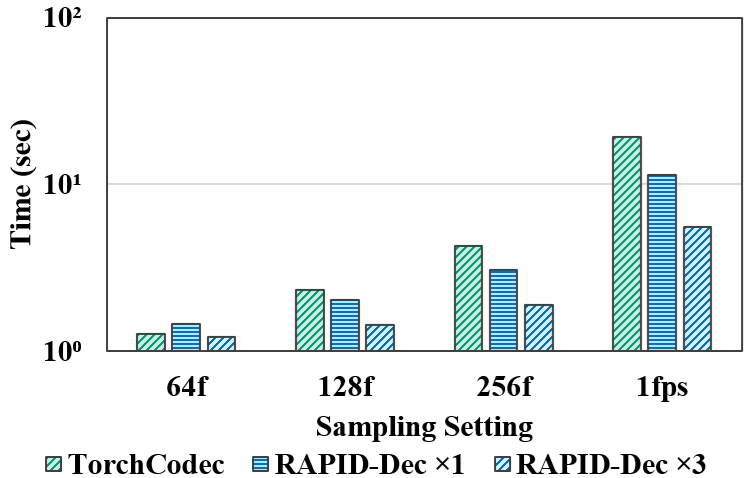}
    \caption{H.265/HEVC}
    \label{fig:nvdec_lat-b}
  \end{subfigure}
  \hfill
  \begin{subfigure}{0.32\linewidth}
    \includegraphics[width=\linewidth]{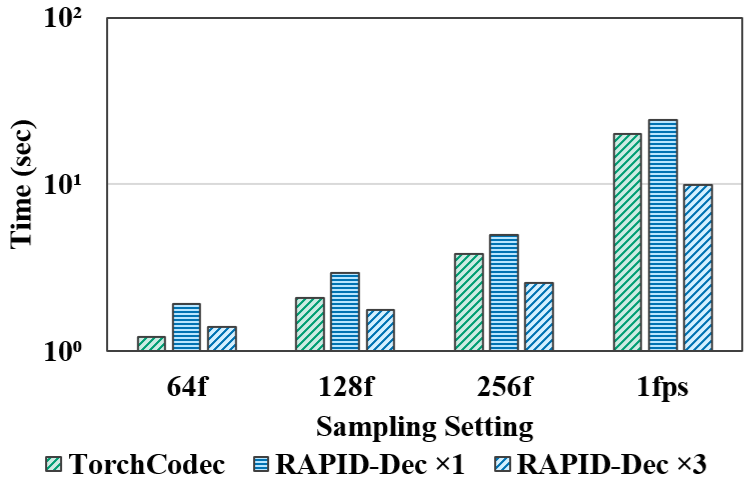}
    \caption{VP9}
    \label{fig:nvdec_lat-c}
  \end{subfigure}
  \vspace{-8pt}
  \caption{Video decoding latency using NVDEC. The y-axis is in log scale.}
  \label{fig:nvdec_lat}
\end{figure*}

\medskip
\noindent\textbf{NVDEC results.}
Fig.~\ref{fig:nvdec_lat} shows the video decoding latency on NVDEC, measured on an RTX 5880 Ada GPU with three NVDEC engines.
Among the baselines, TorchCodec~\citep{torchcodec} is the only library whose official release supports NVDEC decoding, so we compare RAPID-Dec with TorchCodec in this setting. QuickCodec~\citep{schneider2026quickvideo} decodes only on the CPU, and the PyPI release of Decord~\citep{decord2019} is built without CUDA.
TorchCodec decodes a video with a single NVDEC decoder, namely one decoding session.
Since a single decoding session is limited to the throughput of one NVDEC engine, TorchCodec effectively uses only one of the three engines.
In contrast, RAPID-Dec partitions the decoding work into groups that each start at a keyframe.
Since these groups can be decoded independently, RAPID-Dec can distribute them across multiple decoders, each running its own decoding session.
We therefore evaluate RAPID-Dec with one decoder (RAPID-Dec~$\times1$) and with three decoders running in parallel (RAPID-Dec~$\times3$) to utilize all three NVDEC engines.

Since RAPID-Dec skips non-reference frames, RAPID-Dec~$\times1$ outperforms TorchCodec on H.264/AVC and H.265/HEVC in all settings except H.265/HEVC at 64 frames, and RAPID-Dec~$\times3$ is the fastest in every H.264/AVC and H.265/HEVC setting, with up to a 4.9$\times$ speedup over TorchCodec.
As explained in Sec.~\ref{sec:codec-impl}, RAPID-Dec treats every VP9 frame as a reference frame, so non-reference skipping does not apply to VP9.
Without this benefit, the CPU-side overhead of RAPID-Dec for indexing and packet handling makes RAPID-Dec~$\times1$ slower than TorchCodec on VP9.
The same overhead affects H.265/HEVC at 64 frames, where little decoding is required.
With three decoders, parallel decoding outweighs this overhead, and RAPID-Dec~$\times3$ outperforms TorchCodec on VP9 from 128 frames onward.
Thus, with all three engines in use, RAPID-Dec reduces the decoding latency on NVDEC in every setting except VP9 at 64 frames.

\begin{figure*}[t]
  \centering
  \begin{subfigure}{0.32\linewidth}
    \includegraphics[width=\linewidth]{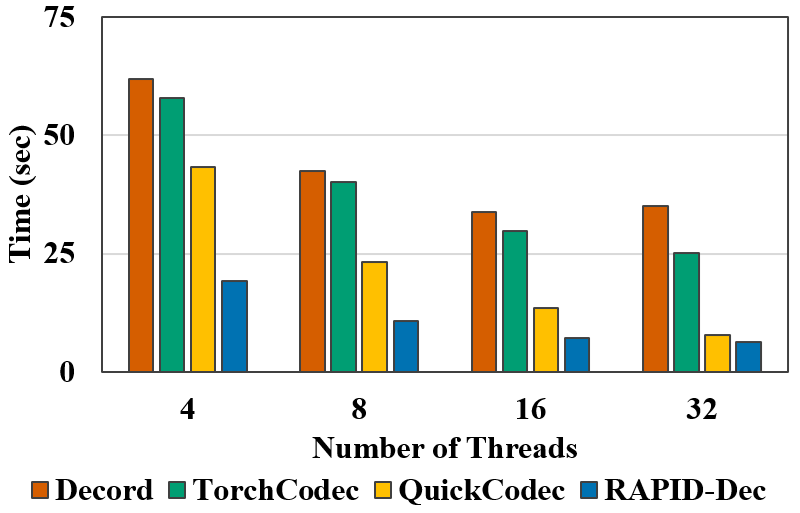}
    \caption{H.264/AVC}
    \label{fig:parallel_lat_amd-a}
  \end{subfigure}
  \hfill
  \begin{subfigure}{0.32\linewidth}
    \includegraphics[width=\linewidth]{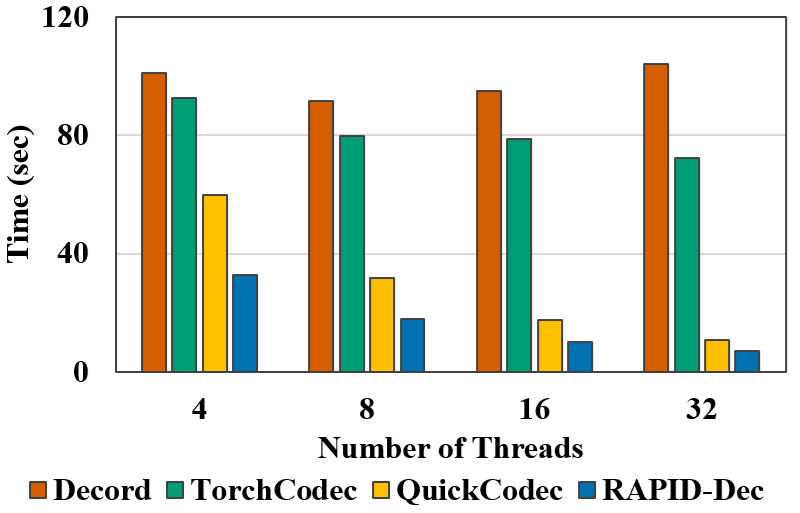}
    \caption{H.265/HEVC}
    \label{fig:parallel_lat_amd-b}
  \end{subfigure}
  \hfill
  \begin{subfigure}{0.32\linewidth}
    \includegraphics[width=\linewidth]{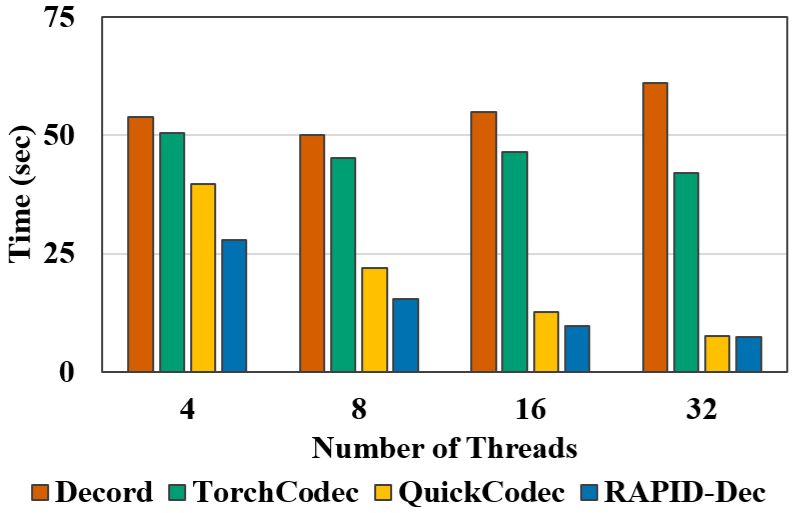}
    \caption{VP9}
    \label{fig:parallel_lat_amd-c}
  \end{subfigure}
  \vspace{-8pt}
  \caption{Video decoding latency of 1~fps sampling (up to 2,048 frames) with different numbers of threads on the AMD CPU.}
  \label{fig:parallel_lat_amd}
\end{figure*}

\begin{figure*}[t]
  \centering
  \begin{subfigure}{0.32\linewidth}
    \includegraphics[width=\linewidth]{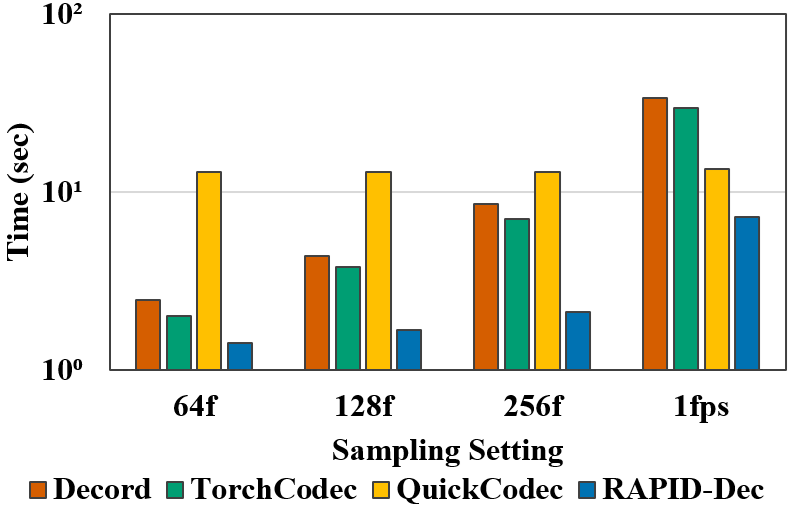}
    \caption{H.264/AVC}
    \label{fig:sample_lat_amd-a}
  \end{subfigure}
  \hfill
  \begin{subfigure}{0.32\linewidth}
    \includegraphics[width=\linewidth]{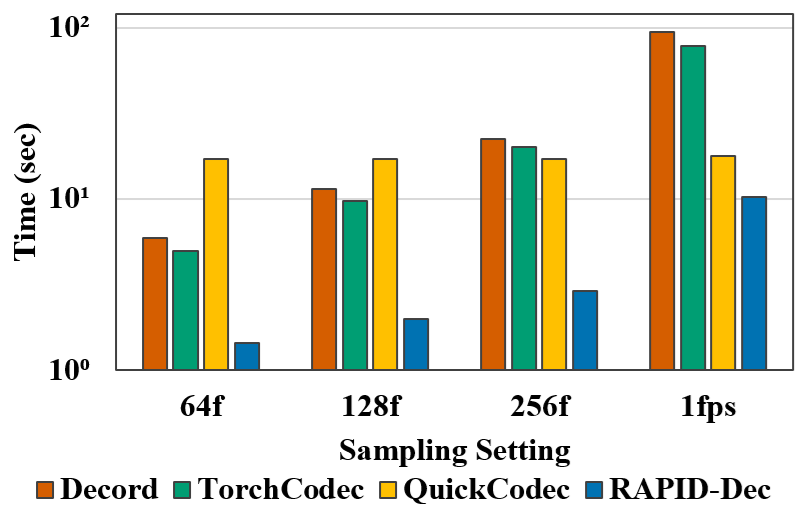}
    \caption{H.265/HEVC}
    \label{fig:sample_lat_amd-b}
  \end{subfigure}
  \hfill
  \begin{subfigure}{0.32\linewidth}
    \includegraphics[width=\linewidth]{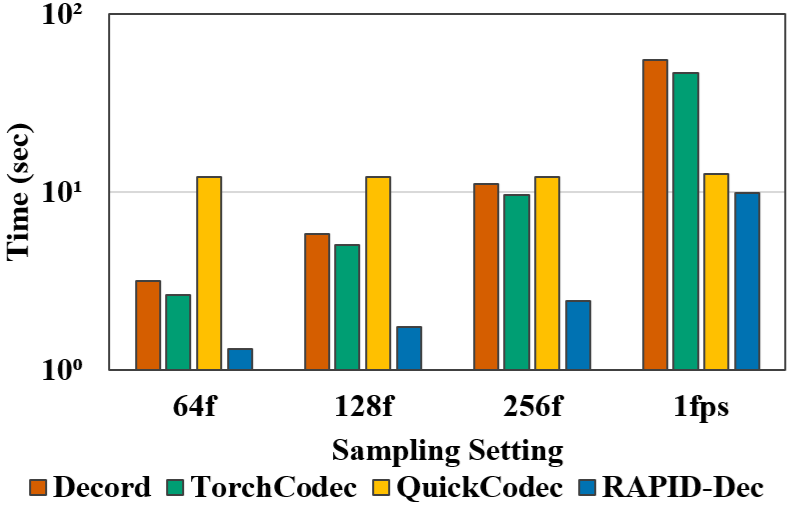}
    \caption{VP9}
    \label{fig:sample_lat_amd-c}
  \end{subfigure}
  \vspace{-8pt}
  \caption{Video decoding latency under different frame sampling settings with 16 threads on the AMD CPU. The y-axis is in log scale.}
  \label{fig:sample_lat_amd}
\end{figure*}
\medskip
\noindent\textbf{Different CPU results.}
In addition to the Intel Xeon Gold 6530 CPU used in the main paper, we evaluate RAPID-Dec on an AMD EPYC 7543 32-core CPU at 2.8\,GHz, with all other settings of Fig.~\ref{fig:parallel_lat} and Fig.~\ref{fig:sample_lat} unchanged.
As shown in Fig.~\ref{fig:parallel_lat_amd} and Fig.~\ref{fig:sample_lat_amd}, the results follow the same trends as on the Intel CPU.
RAPID-Dec achieves the lowest latency in every setting, Decord and TorchCodec scale poorly with the number of threads, and the latency of QuickCodec remains nearly constant across sampling settings.
With 1~fps sampling and the default 16 threads, RAPID-Dec is 1.9$\times$, 1.7$\times$, and 1.3$\times$ faster than QuickCodec, the fastest baseline, on H.264/AVC, H.265/HEVC, and VP9, respectively, close to the 2.0$\times$, 1.7$\times$, and 1.4$\times$ speedups on the Intel CPU.
Under the other sampling settings with 16 threads, RAPID-Dec is 1.4--3.5$\times$, 2.3--4.9$\times$, and 3.3--5.9$\times$ faster than the fastest baseline in each setting at 64, 128, and 256 frames across the three codecs, respectively.

\begin{figure*}[t]
  \centering
  \begin{subfigure}{0.32\linewidth}
    \includegraphics[width=\linewidth]{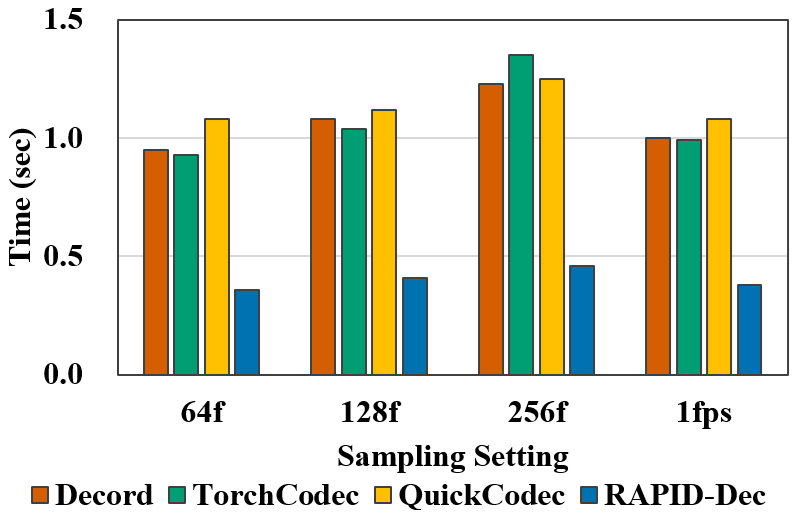}
    \caption{Short ($<$2 min)}
    \label{fig:vme_lat-short}
  \end{subfigure}
  \hfill
  \begin{subfigure}{0.32\linewidth}
    \includegraphics[width=\linewidth]{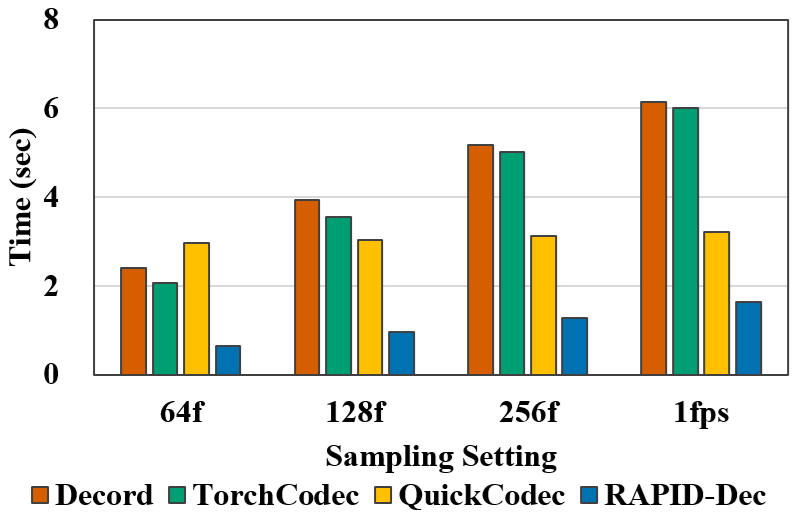}
    \caption{Medium (4--15 min)}
    \label{fig:vme_lat-med}
  \end{subfigure}
  \hfill
  \begin{subfigure}{0.32\linewidth}
    \includegraphics[width=\linewidth]{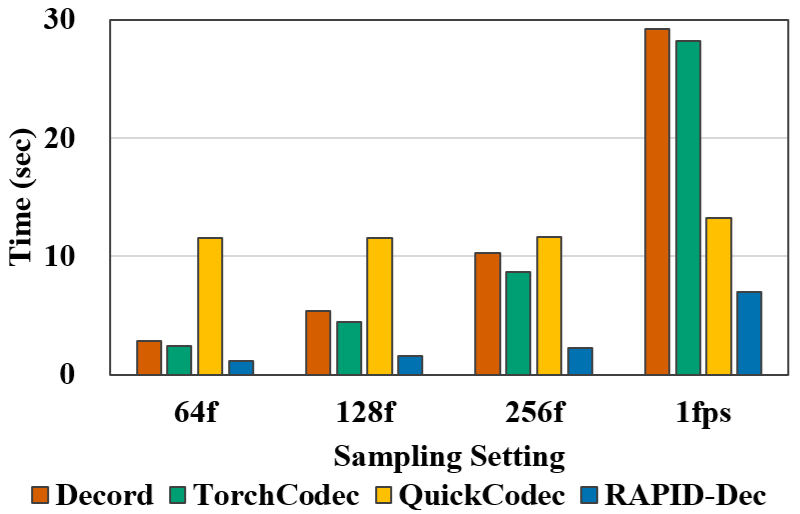}
    \caption{Long (30--60 min)}
    \label{fig:vme_lat-long}
  \end{subfigure}
  \vspace{-8pt}
  \caption{Video decoding latency on the Video-MME videos of each duration with 16 threads.}
  \label{fig:vme_lat}
\end{figure*}
\begin{figure*}[t]
  \centering
  \begin{subfigure}{0.32\linewidth}
    \includegraphics[width=\linewidth]{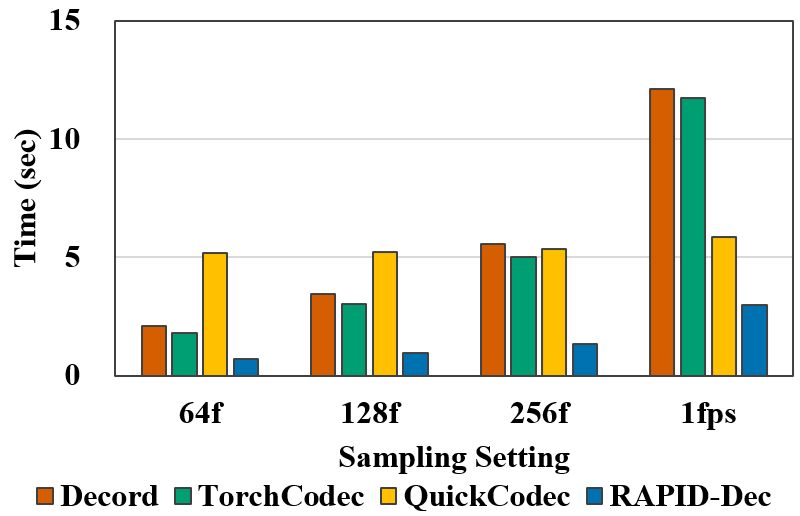}
    \caption{All videos}
    \label{fig:vme_lat2-avg}
  \end{subfigure}
  \hfill
  \begin{subfigure}{0.32\linewidth}
    \includegraphics[width=\linewidth]{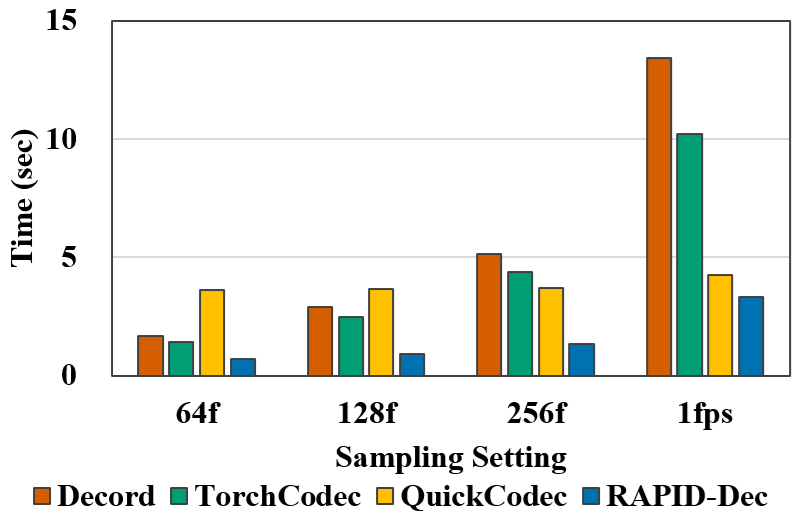}
    \caption{No non-reference frames}
    \label{fig:vme_lat2-ref}
  \end{subfigure}
  \hfill
  \begin{subfigure}{0.32\linewidth}
    \includegraphics[width=\linewidth]{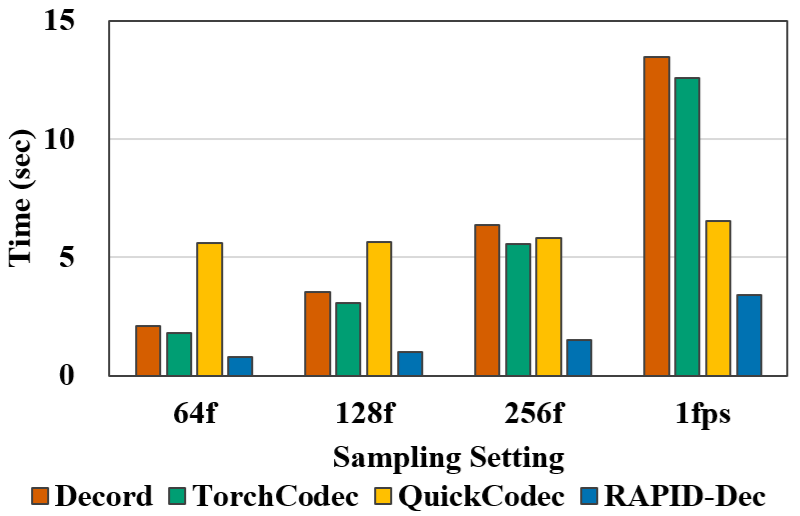}
    \caption{Long group lengths}
    \label{fig:vme_lat2-gop}
  \end{subfigure}
  \vspace{-8pt}
  \caption{Video decoding latency on all Video-MME videos and on two special cases with 16 threads.}
  \label{fig:vme_lat2}
\end{figure*}

\subsection{Large-scale evaluation of RAPID-Dec}
To examine whether the speedup of RAPID-Dec generalizes beyond the four hour-long videos, we repeat the experiment from Fig.~\ref{fig:sample_lat} on all 900 videos of Video-MME, all of which are encoded in H.264/AVC, with the same number of threads and sampling settings. Since the results are averaged over many videos, we decode each video once.

\medskip
\noindent\textbf{Impact of video duration.}
Fig.~\ref{fig:vme_lat} shows the latency for short, medium, and long videos, whose median durations are about 1, 8, and 40 minutes, respectively.
RAPID-Dec achieves the lowest latency in every sampling setting for all three durations.
Since QuickCodec decodes every frame, its gap to RAPID-Dec widens as sampling becomes sparser, most notably on long videos, which contain the most frames.
In contrast, the speedup of RAPID-Dec over TorchCodec grows from 64 to 256 frames on medium and long videos, since more groups contain targets and RAPID-Dec decodes them in parallel.
As a result, the fastest baseline changes with the sampling setting and the video duration, whereas RAPID-Dec is the fastest in every case.
Averaged over all videos (Fig.~\ref{fig:vme_lat2-avg}), RAPID-Dec is 1.9--3.7$\times$ faster than the fastest baseline in each sampling setting, close to the 1.7--3.7$\times$ on the hour-long H.264/AVC videos in Fig.~\ref{fig:sample_lat-a}.

\medskip
\noindent\textbf{Special cases.}
We further examine two subsets of videos that may limit RAPID-Dec.
The first subset consists of the videos without non-reference frames, which account for 4.4\% of the videos and for which non-reference skipping has no effect.
As shown in Fig.~\ref{fig:vme_lat2-ref}, the speedup of RAPID-Dec on these videos is smaller than on all videos, especially at 1~fps, where non-reference skipping is the most effective component (Table~\ref{tab:dec_ablation}).
Nevertheless, RAPID-Dec still achieves the lowest latency in every setting.
The second subset consists of the 5\% of videos with the longest median group length, above 5.75~s.
Longer groups leave fewer groups to decode in parallel.
However, encoders insert a keyframe whenever a group reaches its maximum length~\cite{x264,x265}, so even these videos are divided into multiple groups, and when only a few groups contain targets, RAPID-Dec uses the remaining threads inside each decoder as explained in Sec.~\ref{sec:rapid-dec}.
Accordingly, the speedup of RAPID-Dec on these videos remains close to that on all videos (Fig.~\ref{fig:vme_lat2-gop}).

\subsection{Candidate overhead of RAPID-Sel}

In addition to the accuracy in Table~\ref{tab:cand-uniform}, Table~\ref{tab:supp-cand} reports the preprocessing latency of RAPID-Sel with uniform and keyframe candidates, which are equal in number for each video.
The latency is measured as in Table~\ref{tab:main} on 10\% of the Video-MME videos, and None denotes uniform sampling without candidates.
Keyframe candidates achieve slightly higher average accuracy, with the largest gain on long videos (+1.56 points), and make preprocessing 2.1$\times$ faster than uniform candidates.
Over uniform sampling without candidates, keyframe candidates add only 0.64\,s, whereas uniform candidates add 2.12\,s.
Since both settings score the same number of candidates, this gap comes mainly from candidate decoding.
A keyframe references no other frame and is decoded on its own, whereas an evenly spaced candidate generally requires decoding the frames it references, starting from the preceding keyframe.
Keyframes are therefore more efficient candidates than evenly spaced frames, while achieving slightly higher accuracy, especially on long videos.
\begin{table}[t]
  \centering
  \caption{Accuracy (\%) and preprocessing latency of uniform vs.\ keyframe candidates of RAPID-Sel on Video-MME with Qwen3-VL-8B and 64 frames.}
  \small
  \setlength{\tabcolsep}{3pt}
  \begin{tabular}{l c cccc}
    \toprule
    Candidates & Preproc.\ (s)$\downarrow$ & Short & Medium & Long & Avg. \\
    \midrule
    None & 0.75 & 74.56 & 61.89 & 54.44 & 63.63 \\
    \midrule
    Uniform   & 2.87 & \textbf{76.22} & 63.67 & 55.22 & 65.04 \\
    Keyframes & \textbf{1.39} & 75.67 & \textbf{63.78} & \textbf{56.78} & \textbf{65.41} \\
    \bottomrule
  \end{tabular}
  \label{tab:supp-cand}
\end{table}

\begin{figure}[t]
  \centering
  \includegraphics[width=\linewidth]{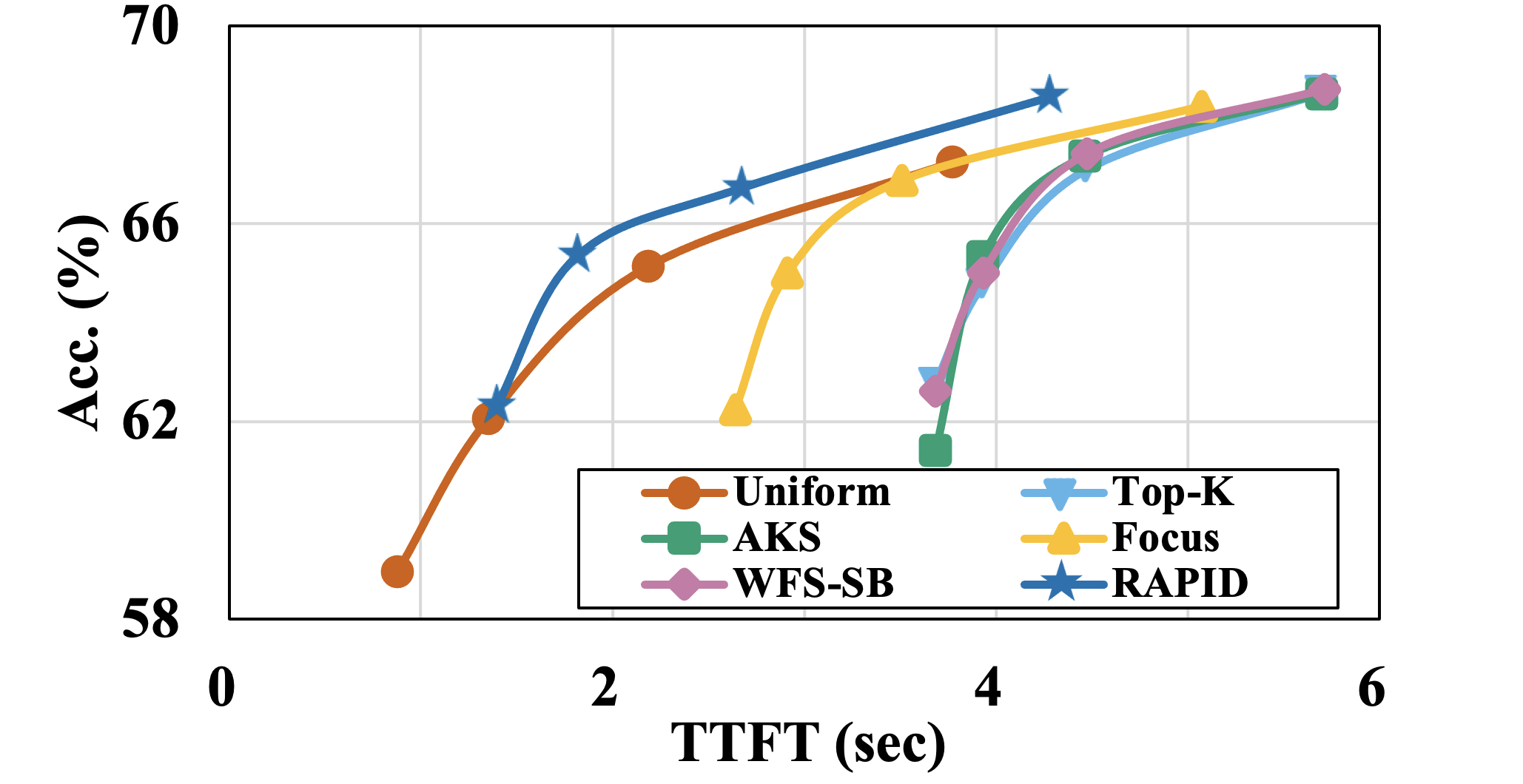}
  \vspace{-18pt}
  \caption{Average accuracy over the three benchmarks vs.\ TTFT of Qwen3-VL-8B with 32 to 256 frames.}
  \label{fig:ttft_rapid}
\end{figure}

\subsection{Accuracy vs.\ TTFT}
Table~\ref{tab:main} compares frame selection methods by preprocessing latency at the same frame budget.
However, the TTFT also depends on the number of input frames, since the VLM encodes and prefills more visual tokens with more frames.
A method with a smaller selection overhead can therefore feed more frames to the VLM within the same TTFT, and uniform sampling has no selection overhead.
To compare the methods under the full pipeline, Fig.~\ref{fig:ttft_rapid} plots the accuracy of Qwen3-VL-8B against its TTFT with 32, 64, 128, and 256 frames, where the points of each method are ordered by the number of frames from left to right.
The accuracy is averaged over Video-MME, MLVU, and LongVideoBench, and the TTFT is measured on the same 10\% of the videos as in Table~\ref{tab:main}, with RAPID-Dec for all methods.
At matched accuracy, RAPID lies to the left of the other methods, indicating that it yields the same accuracy with a lower TTFT.
At matched TTFT, RAPID lies above the other methods, achieving higher accuracy.
Both trends hold over the entire range covered by RAPID, except with 32 frames, where RAPID performs on par with uniform sampling with 64 frames.
The other selection methods reach accuracy similar to that of RAPID, but their heavy selection overhead makes their TTFT higher than that of RAPID at the same number of frames.
RAPID thus provides a better accuracy--TTFT trade-off than both uniform sampling with more frames and existing frame selection methods.

\section{Analysis}\label{sec:appx-analysis}
\begin{figure}[t]
  \centering
  \includegraphics[width=\linewidth]{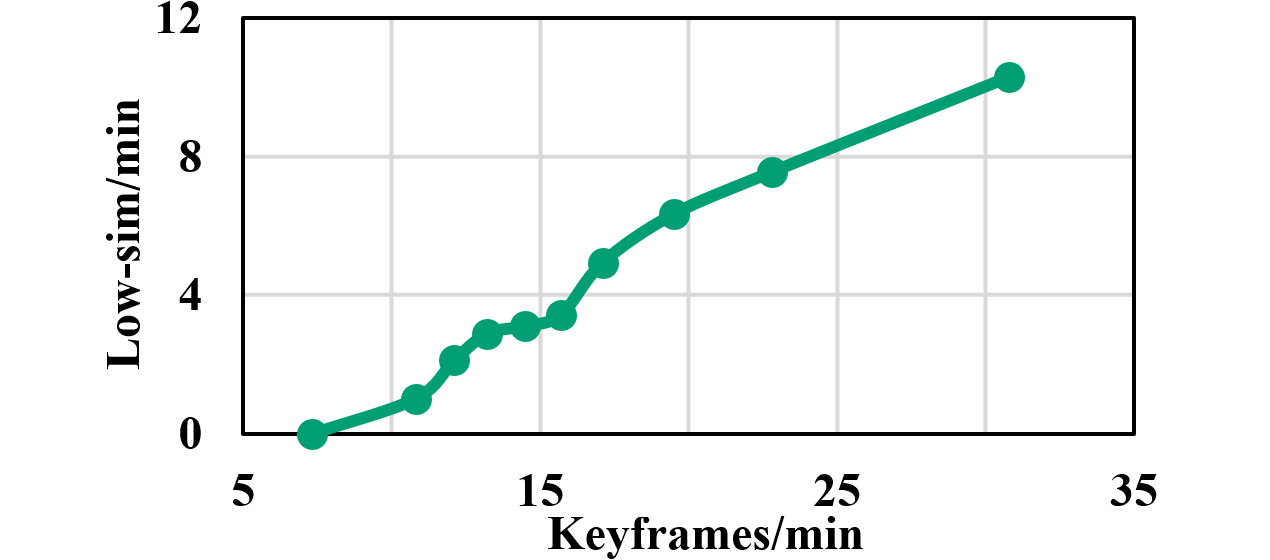}
  \vspace{-18pt}
  \caption{Medians of keyframes and low-similarity intervals (Low-sim) per minute in ten groups of videos sorted by keyframes per minute.}
  \label{fig:kf-density}
\end{figure}
\subsection{Keyframes and visual changes}
As shown in Fig.~\ref{fig:scene-key-a}, the median per-video fraction of low-similarity intervals containing a keyframe is 0.81 over the videos with at least five such intervals, so that each fraction is computed from enough intervals.
If keyframes were placed regardless of the content, this fraction would be close to the fraction of all 1~s intervals containing a keyframe.
However, the latter has a median of only 0.26 over the same videos, and the actual value (0.81) is $3.1\times$ higher. This further indicates that keyframes are placed according to visual changes rather than independently of the content.

We next examine whether videos with more visual changes contain more keyframes.
For each video, we compute the number of keyframes per minute and the number of low-similarity intervals per minute, using the same threshold of 0.73 as in Sec.~\ref{sec:exp-sel}.
Unlike the fraction, these counts are divided by the video duration rather than by the number of low-similarity intervals, so we also include the videos excluded above.
We sort the videos by keyframes per minute, divide them into ten groups of equal size, and plot the median of each group in Fig.~\ref{fig:kf-density}.
The number of low-similarity intervals per minute increases with the number of keyframes per minute, with a Spearman correlation of 0.61 over individual videos.
This indicates that videos with more visual changes tend to contain more keyframes.

\begin{figure*}
  \centering
  \includegraphics[width=\linewidth]{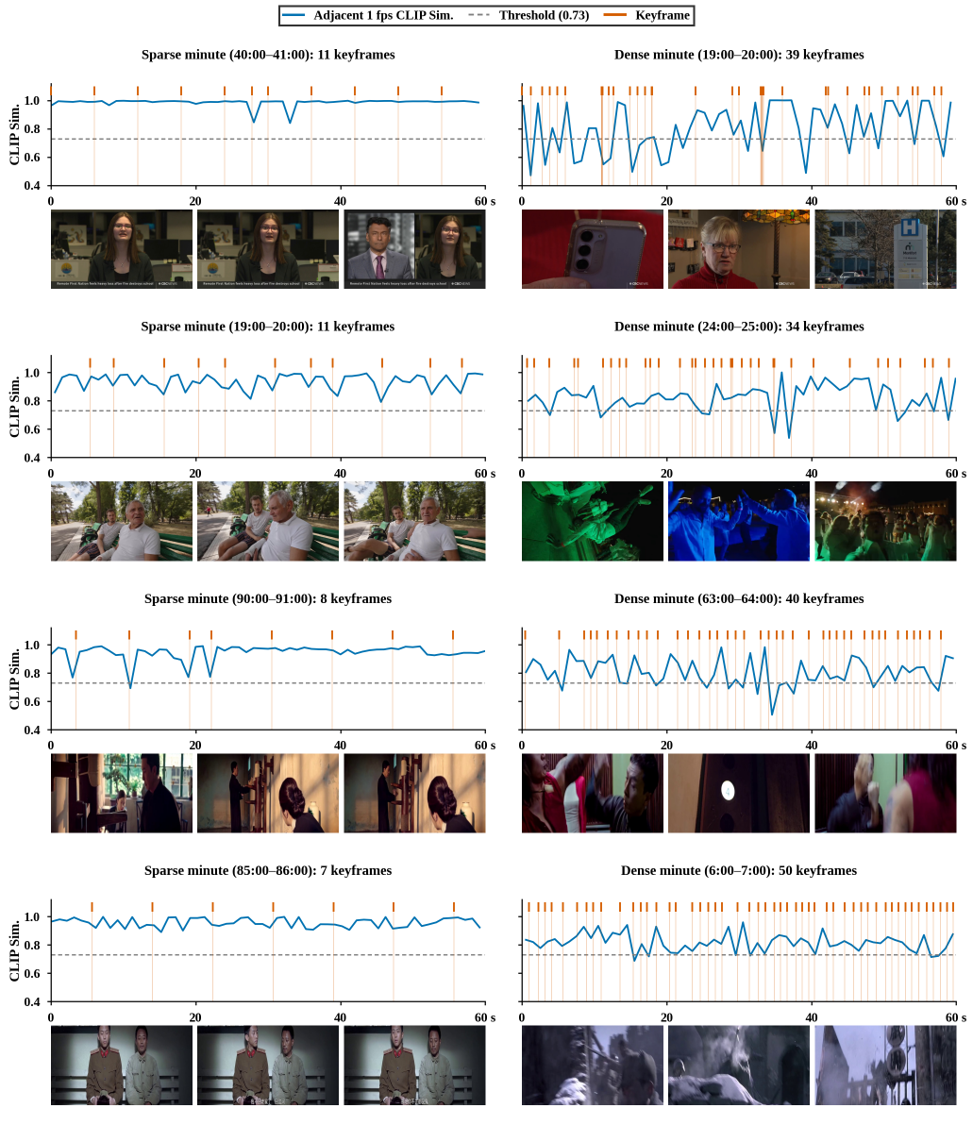}
  \vspace{-18pt}
      \caption{Keyframes in a sparse minute (left) and a dense minute (right) of four videos. Each plot shows the CLIP similarity of adjacent 1~fps frames, the threshold of 0.73 for low-similarity intervals, and the keyframes of the minute. Below each plot are three consecutive keyframes of the minute.}
  \label{fig:kf-examples}
  \vspace{-10pt}
\end{figure*}

\subsection{Examples of keyframe placement}
The analyses above suggest that keyframes are placed at visual changes and that videos with more visual changes contain more keyframes.
Fig.~\ref{fig:kf-examples} illustrates that this also holds within a single video by comparing a sparse minute and a dense minute of four videos.
Since both minutes of a video come from the same encoding, the difference in their numbers of keyframes reflects their content.
In these examples, the dense minute contains 3.1--7.1$\times$ as many keyframes as the sparse minute.
In the sparse minutes, the CLIP similarity stays close to 1 and rarely falls below the threshold of 0.73 used for low-similarity intervals in Fig.~\ref{fig:scene-key}, and consecutive keyframes often show the same shot.
In the dense minutes, the similarity is lower and fluctuates more, and the three consecutive keyframes, only a few seconds apart, show different shots.

The sparse minutes show people sitting and talking with little motion.
In contrast, the dense minutes show rapidly changing content, namely a sequence of short shots, a night party under colored lights, a fight inside an elevator, and a battle scene.
Even in the sparse minutes, keyframes still appear every few seconds.
In the first and last videos, they are placed at nearly regular intervals, likely because encoders insert a keyframe when a group reaches its maximum length, as discussed for the highly similar pairs across a keyframe in Fig.~\ref{fig:scene-key-b}.
Such keyframes do not mark visual changes, but they still represent their groups since the content changes little within each group.
As a result, the keyframes that RAPID-Sel scores are dense in dynamic segments and sparse in static segments, where a few keyframes suffice to represent long groups of similar frames.

\section{Limitations and future work} 
\label{sec:appx-future}
RAPID-Sel represents each group by its keyframe, so it estimates the relevance of a group only at a coarse level.
Visual content that changes gradually within a group, as suggested by the slightly lower similarity between a keyframe and the other frames of its group (Fig.~\ref{fig:scene-key-b}), is not reflected in the relevance score.
RAPID-Sel may therefore miss frames needed for questions about fine-grained actions.
Motion vectors in the bitstream indicate how much the content moves within a group, and combining them with the query relevance of keyframes to select frames is a promising direction for future work.
In addition, our implementation of RAPID-Dec supports H.264/AVC, H.265/HEVC, and VP9.
Newer codecs such as AV1~\citep{av1} and H.266/VVC~\citep{wang2021vvc_hls} also signal keyframes and reference information in their headers, so RAPID-Dec could be extended to them by parsing these headers.

%

\end{document}